\documentclass{article}

\PassOptionsToPackage{numbers, compress}{natbib}

\usepackage[preprint]{nips24}

\usepackage[utf8]{inputenc} 
\usepackage[T1]{fontenc}    
\usepackage{hyperref}       
\usepackage{url}            
\usepackage{booktabs}       
\usepackage{amsfonts}       
\usepackage{nicefrac}       
\usepackage{microtype}      
\usepackage{xcolor}         
\usepackage{algorithm}
\usepackage{algorithmicx}
\usepackage{algpseudocode}

\usepackage{amsmath, amsfonts, amssymb, amsthm}
\usepackage{hyperref}
\usepackage{cleveref}
\usepackage{xcolor}
\usepackage{url}            
\usepackage{booktabs}       
\usepackage{nicefrac}       
\usepackage{microtype}      
\usepackage{graphicx}
\usepackage{caption}
\usepackage{float}
\usepackage{mdframed}
\usepackage{fontawesome5}

\newtheorem{theorem}{Theorem}
    \newtheorem{proposition}{Proposition}
\newtheorem{lemma}{Lemma}

\newtheorem{assumption}{Assumption}
\newtheorem{remark}{Remark}

\usepackage{enumitem}
\newlist{todolist}{itemize}{2}
\setlist[todolist]{label=$\square$}
\usepackage{pifont}
\newcommand{\orange}[1]{\textcolor{orange} {{{#1}}}}

\newcommand{\E}{\mathbb{E}}
\newcommand{\R}{\mathbb{R}}
\newcommand{\N}{\mathbb{N}}
\newcommand{\x}{\mathbf{x}}

\newcommand{\W}{\mathbf{W}}
\newcommand{\mednorm}[1]{\| #1\| }
\newcommand{\trace}[1]{\mathrm{Tr}\left(#1\right)}

\title{
How Neural Networks Learn the Support \\ is an Implicit Regularization Effect of SGD
}

\author{
Pierfrancesco Beneventano\footnotemark[1] \\
Princeton University\\
\texttt{pierb@princeton.edu} \\
\And
Andrea Pinto\thanks{
Equal contributions. PB developed the mathematical proofs. \\
\faGithubSquare \; Reproducible code can be found at 
\url{github.com/andreakiro/sgd-finds-support}.
} \\
Massachusetts Institute of Technology
\\
\texttt{pintoa@mit.edu} \\
\And
Tomaso Poggio \\
Massachusetts Institute of Technology \\
\texttt{tp@csail.mit.edu} \\
}

\begin{document}

\maketitle

\begin{abstract}
We investigate the ability of deep neural networks to identify the support of the target function. Our findings reveal that mini-batch SGD effectively learns the support in the first layer of the network by shrinking to zero the weights associated with irrelevant components of input. In contrast, we demonstrate that while vanilla GD also approximates the target function, it requires an explicit regularization term to learn the support in the first layer. We prove that this property of mini-batch SGD is due to a second-order implicit regularization effect which is proportional to $\eta / b$ (step size / batch size).
Our results are not only another proof that implicit regularization has a significant impact on training optimization dynamics but they also shed light on the structure of the features that are learned by the network. 
Additionally, they suggest that smaller batches enhance feature interpretability and reduce dependency on initialization.
\end{abstract}

\begin{figure}[ht]
\centering
\includegraphics[width=\textwidth]{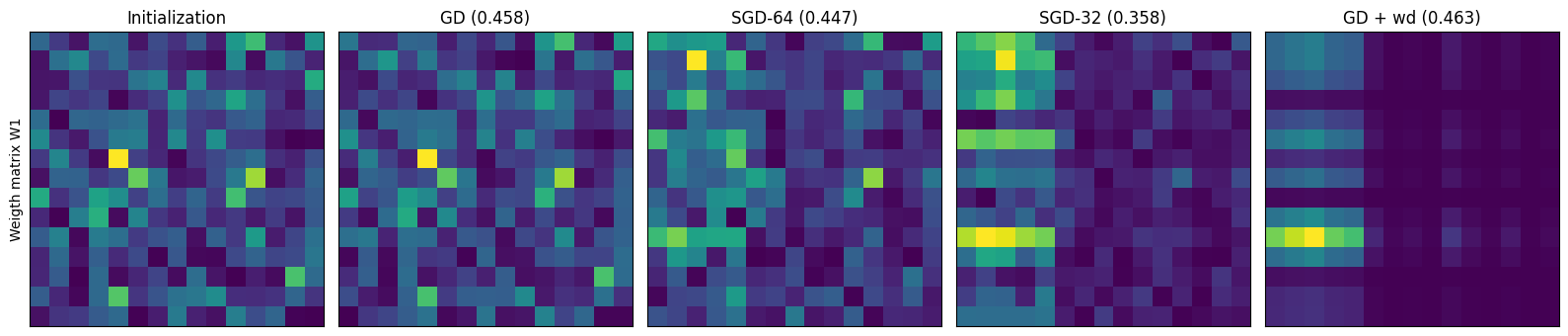}
\includegraphics[width=\textwidth]{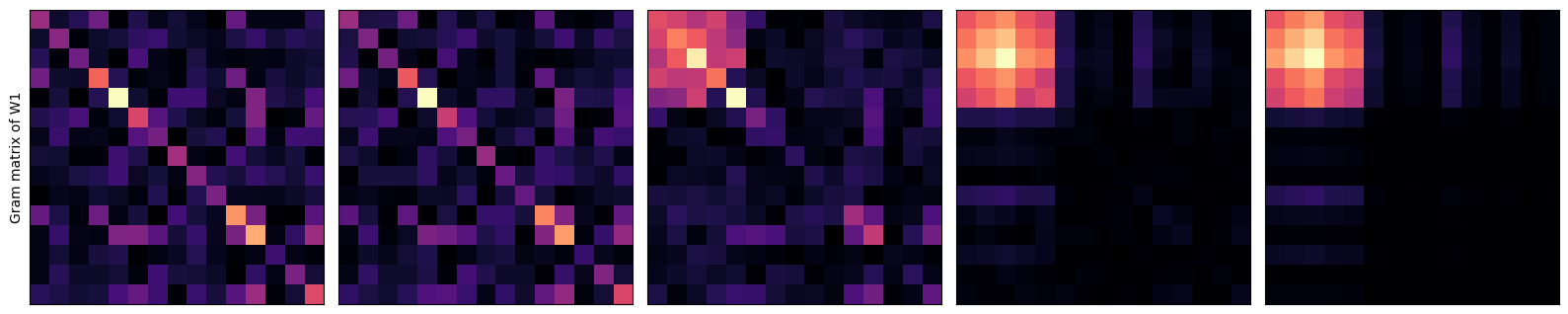}
\caption{
\textbf{First layer weights and their Gram matrices at initialization and at convergence for GD and SGDs.}
All trained from same initialization on target $y(\x) = \sin (\sum_{i<r} x_i )$ with sparse support $r < d$. 
While all network achieve similar performance on loss (in brackets on the right), SGD with smaller batches identifies the support in the first layer weights.
}
\label{fig:abstract:support:sine:gw}
\end{figure}


\section{Introduction}

\subsection{The problem}
Neural networks, like any function approximator, must properly identify the support of the target function -- the subset of input features that significantly influence the output -- to learn effectively.
The convergence of the training loss implies that the support of the target is correctly identified on the training data points. 
However, the vulnerability to adversarial attacks indicates that this identification is often incomplete over the entire domain of the target function and improves with increased model robustness.
We investigate how generalizing neural networks in regression and simple computer vision tasks empirically identify the support, \textit{finding that this typically occurs in the first layer}. 
Contrary to what one might assume, this phenomenon is an artifact of the training procedure, influenced by hyperparameter choices (see Fig.\ \ref{fig:abstract:support:sine:gw}), it is not induced by learning the target function itself on the training set.

\subsection{Our contribution} 

Our contribution is twofold. One one hand we are the first to our knowledge to identify \textit{what the role of the first layer of neural networks is or, analogously, where neural networks identify the support of the represented function}. This has implications on the interpretability of neural networks and on our ability to quantify and understand in what way neural networks depend on the initialization after a given training time, see \S \ref{section:implications}. 
On the theory side, we are able to mathematically describe the dynamics of the weights of (linear) neural networks interacting with degenerate directions for the target function.
Indeed, the community is aware that the target functions in practice are usually representable as lower-dimensional function and real-world neural networks have Hessians and weight matrices which are extremely degenerate. However, in the very rich literature dealing with the convergence of training procedures for neural networks, most works could only technically deal with the case in which \textit{both} the weight matrices \textit{and} the linear target function are full rank, see, e.g., \cite{arora_convergence_2019,arora_implicit_2019}. What we believe is an important theoretical novelty of this work is also that while all the works on linear networks deal with a approximately balanced case, see, e.g., \cite[Definition 1]{arora_convergence_2019}, this is the first time to our knowledge in which someone proves that even initializing with arbitrarily unbalanced weights not only we obtain converge of the loss, but \textit{converge to a balanced solution without any explicit regularization}.

\paragraph{Theoretical Contributions.}
We prove in Theorem \ref{theo:shrinking} that both full-batch GD and mini-batch SGD training dynamics shrink the weights of the first layer of linear networks corresponding to irrelevant input features. This phenomenon occurs faster for mini-batch SGD due to a second-order implicit regularization effect that is linearly proportional to the learning rate $\eta$ divided by the batch size $b$.
We proceed to show in Proposition \ref{prop:GD} that in the case of diagonal linear or ReLU networks, full-batch GD learns the target's support, however it happens uniformly throughout the layers, often not in the first one.
In Proposition \ref{prop:SGD}, we show instead that mini-batch SGD often succeeds in learning the support of the target function in the first layer, in the case of diagonal linear networks and we specify the required training time in Theorem \ref{theo:time}. Precisely, this happens whenever the trajectory presents oscillations, such as in cases of noisy labels or misspecified data.

As we explain in \S\ref{section:implicit_reg}, learning the support of the target function happens in two different phases of the training dynamics. A first phase in which the loss converges and the network learns the support, followed by a second phase where the implicit regularization effect of SGD aligns the first layer with it.
We conclude by discussing in \cref{section:ReLU} how these results for linear networks generalize to ReLU networks and commenting on the fact that the weights interacting with the irrelevant components that are not spanned by the data do converge to zero if and only if weight decay is added.

\paragraph{Empirical experiments and appendix.} Our claims are supported by extensive empirical evidence on several synthetic target functions we crafted to have a clear handle on the irrelevant components of input. We performed ablations on MLP (width \& depth), activations, parameter initialization (several seeds), learning rates, batch sizes, and sampled random SGD trajectories (See Appendix \ref{appendix:support:synth}). While this set of experiment is already very indicative of the optimization dynamics we claim, we also performed ablations on vision datasets to test our findings in more practical scenarios. We present similar results on vanilla MLP trained on MNIST digits \cite{lecun1998mnist} in the main text and on fine-tuning an MLP head on top of a pretrained ResNet16 \cite{he2015deep} on CIFAR10 dataset \cite{krizhevsky2009learning} in Appendix \ref{appendix:support:cifar10}.

\subsection{Background}
\paragraph{The targets are inherently low-dimensional.}
It is empirically well-known that the signal in the data is often inherently low-dimensional or can be represented as such. In particular, in Appendix \ref{section:low_dimensional} we show that in the case of MNIST and CIFAR10 most of the signal lies in a specific (non-linear) 9-dimensional space, though the input covariance is high dimensional, as it has many sizable singular values. This 9-dimensional non-linear space is obtained by processing the input through a trained CNN.
On top of this, it has been proved that all the functions that are efficiently Turing Computable (that is computable in less than exponential time in the input dimension), admit a sparse compositional computational graph structure where each node is a function with bounded, "small" input dimension \cite{poggiofraser2024}. Furthermore, they can be approximated without curse of dimensionality by an appropriately deep and sparse network\footnote{The assumption in several recent papers (e.g. see \cite{10.1214/19-AOS1875,kohler2020rate} of a hierarchical composition structure of the regression function can be replaced by its efficient Turing computability.} This implies that target functions that are learnable may often have a small number $r$ of relevant input components\footnote{They certainly have a small number of input components for each of the constituent functions that depend directly on the input variables.}. 

\paragraph{Finding the support.} 
For linear regression it is clear that regularized GD and SGD find the correct support, independently of initialization, but the unregularized versions do not \cite{Memo144}.
Since most of the SOTA neural networks architectures have MLP blocks, it is important to know whether MLPs can recover this low dimensional structures and what is the right training procedure to do so.
Lately, many works study whether neural networks are learning the support in the case of \textit{infinite data} coming from particular distributions, in particular situations as parities, XOR, or regression with single index models, and, often, trained with particular ad-hoc training procedures
\cite{abbe_provable_2023,cornacchia_mathematical_2024,abbe_provable_2019,abbe_poly-time_2020,abbe_merged-staircase_2022,abbe_sgd_2023,damian_neural_2022,damian_optimal_2023,bietti_learning_2022}. 
We tackle this problem from the optimization perspective on the general practical case of a finite dataset with no assumptions on the target function, weak assumptions on the input data, and for the most common training procedures as GD and mini-batch SGD.

\begin{figure}[ht]
\centering
\includegraphics[width=1.0\textwidth]{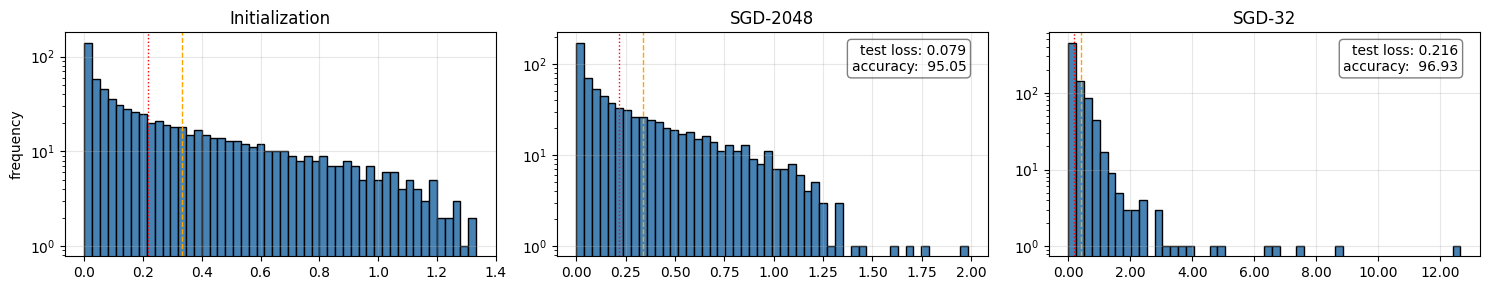}
\caption{\textbf{Eigenvalues histogram for the weights $\W_1$ of the first layer on MNIST dataset.}}
\label{fig:text:vision:mnist:hist}
\end{figure}

\paragraph{SGD implicitly induces regularization.}
Gradient descent (GD) and mini-batch stochastic gradient descent (SGD) are cornerstone techniques in the optimization of neural networks. While GD updates the model parameters by computing the gradient of the loss function across the entire dataset, SGD optimizes using subsets of the data at time, known as mini-batches. This approach enhances computational efficiency and is often more effective in various practical scenarios \cite{bottou_stochastic_2012, lecun_efficient_2012, bengio_practical_2012}. The loss landscapes for neural networks are complex and high-dimensional manifolds dotted with saddles and frequent spurious minima, characterized by highly degenerate Hessians \cite{liu_toward_2020, nguyen_connected_2019, zhang_embedding_2022}. Many of these stationary points generalize poorly to new data \cite{neyshabur_search_2015}. However, an interesting feature of SGD is its implicit regularization effect, which, even in the absence of explicit regularization terms, tends to favor flatter minima or minima with smaller Fisher information matrices \cite{keskar_large-batch_2016, jastrzebski_catastrophic_2021, beneventano_trajectories_2023}. Moreover, these effects are shown to be stronger as $\eta / b$ (step size / batch size) grows \cite{goyal_accurate_2017,jastrzkebski2017three,he_control_2019,beneventano_trajectories_2023}. These minima are known to generally exhibit better generalization to new data \cite{gunasekar_implicit_2017, neyshabur_exploring_2017}.

These observations have inspired numerous theoretical studies on \emph{implicit regularization} to find out which minima are favored by a range of optimization algorithms and given choices of hyperparameters, and our work goes in this direction. A long line of work examines the convergence behaviors of gradient flow and gradient descent for various models such as (diagonal) linear networks with balanced initialization \cite{arora_convergence_2019,arora_implicit_2019,gidel_implicit_2019,vaskevicius_implicit_2019,woodworth_kernel_2020}. Another important line of work investigates the effects of injecting independent noise in the optimization steps, showing that certain noise patterns can direct the convergence to specific types of stationary points \cite{damian_label_2021,li_what_2022,chen_stochastic_2023,orvieto_explicit_2023,shalova_singular-limit_2024}.  However, most of these findings do not apply to mini-batch SGD \cite{yaida_fluctuation-dissipation_2018, li_validity_2021,smith_origin_2021}, and these stochastic algorithms typically converge to qualitatively different minima \cite{haochen_shape_2020}. Recent studies \cite{smith_origin_2021,roberts_sgd_2021} have shown that SGD without sample replacement, that is the standard of practitioners using libraries like PyTorch \cite{paszke2019pytorch} and TensorFlow \cite{tensorflow2015-whitepaper}, exhibits a particular implicit bias. Building on these seminal works, \cite{beneventano_trajectories_2023} proved that SGD without replacement in expectation follows GD along the directions of high curvature, while simultaneously regularizing the trace of the covariance of the gradients along the flatter directions. 
However, these studies all focus on path-dependent regularization in generic settings and do not yet provide convergence results nor specifically address neural networks. We believe we partially bridged this gap by offering insights on how GD and SGD impact the first layer of the trained network.


\section{Identification of the support in the first layer}

Suppose we are given a dataset $\mathcal{D} = \left\{ \x_i, y(\x_i) : \x_i \in \mathbb{R}^d \right\}_{i=1}^n$ with the target function $y(\mathcal{X})$. Denote by $\E_{\mathcal{D}}$ the empirical expectations (averages) over the dataset. Denote the $j$ component of a vector $x$ as $x[j]$. Let us call $r$ the dimensionality of the signal, that is the rank of $\E_{\mathcal{D}}\left[ y(\x) \x^\top \right]$.
Let us call its $r$ right-eigenvectors of the non-zero eigenvalues \emph{relevant} features, which we refer to as the support of the target. Let us call \emph{irrelevant} input features the vectors in the kernel $\E_{\mathcal{D}}\left[ y(\x) \x^\top \right]$.
\begin{assumption}
\label{ass:1}
    Let $v$ be a relevant and $w$ an irrelevant directions in the input space. We assume that the expectation over the dataset of relevant times irrelevant features  $\E_{\mathcal{D}}[ \langle v, \x\rangle \cdot \langle w, \x\rangle]=0$ is zero.
\end{assumption}
This assumption corresponds to claiming that there exists an orthonormal ordered basis of right eigenvectors of $\E_{\mathcal{D}}\left[ y(\x) \x^\top \right]$ for which the matrix $\E_{\mathcal{D}}[\x \x^\top]$ is represented as a block matrix of the form 
$\big(\substack{\Lambda_r 0 \\ 0 \Lambda_i}\big)$ with $\Lambda_i \in \R^{(d-r) \times (d-r)}$ diagonal.
We will work in this basis WLOG from now on.
We denote by $u$ the number of irrelevant input features which are not spanned by the data, i.e., they are in the kernel of $D$. An example of these are the corner pixels of MNIST dataset which are white (valued zero) for every data point. We denote by $i$ the number of the remaining irrelevant features, for which $\E_{\mathcal{D}}\big[\x[j]^2\big] > 0$. Thus $d = r + i + u$.

\begin{figure}[ht]
\centering
\includegraphics[width=0.49\textwidth]{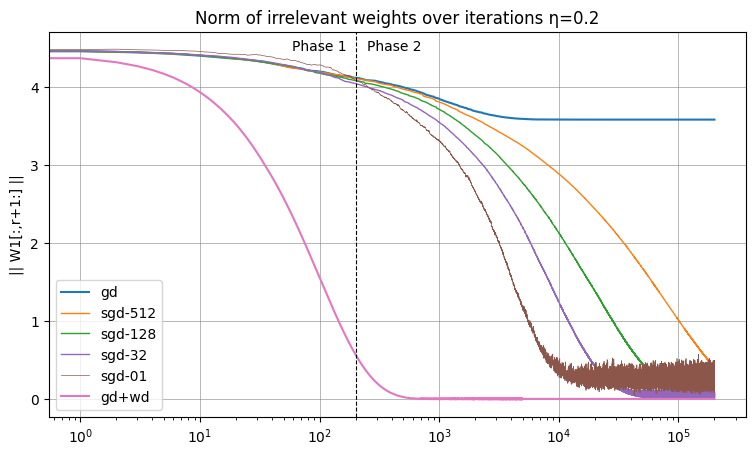}
\includegraphics[width=0.49\textwidth]{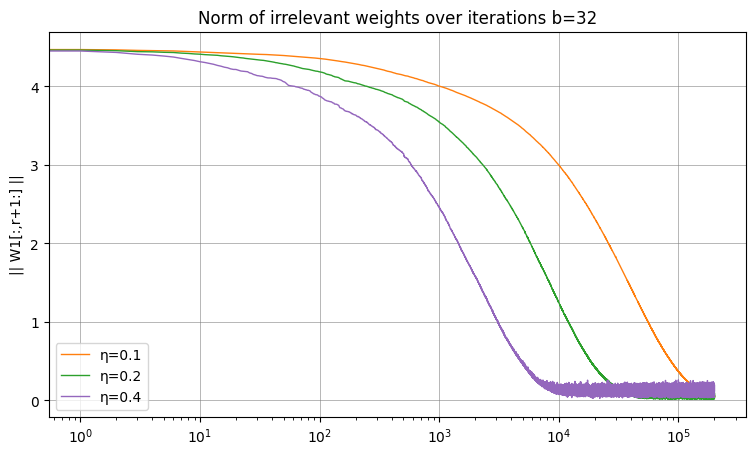}
\caption{\textbf{Norm of irrelevant weights of $\W_1$ over time in a linear network.}}
\label{fig:text:support:linear:irelnorms}
\end{figure}

We deal in this section with deep fully connected neural networks equipped with linear activations and MSE-loss. We prove that \textit{every step} of the optimization dynamics is biased towards networks where the first layer selects the support of the learned features. Precisely, we prove that the weights associated with irrelevant components of the input are progressively annealed by both GD and mini-batch SGD. 
More importantly, our result implies that this phenomenon occurs more rapidly with smaller mini-batches.
We define $f_\theta(x) = \W_L \cdot \sigma \circ \W_{L-1} \cdot \ldots \cdot \sigma \circ \W_1 \cdot x$ to be our neural network with $i$th weight matrix $W_i$ and activation function $\sigma$ and depth $L$.
\begin{theorem}
\label{theo:shrinking}
Let $\sigma$ be the identity, the loss be MSE, Assumption \ref{ass:1} hold.
For every entry $i,j$ of the weights of the first layer corresponding to the irrelevant components $\W_1[:, r+1:]$ there exists $a,c \geq 0$ such that one step of GD or of mini-batch SGD with batch size $b \in \N$, in expectation over batch sampling, are equivalent to multiplying this weight by
\begin{equation*}
\begin{split}
\W_1[i, j] \quad \curvearrowleft \quad
\begin{cases}
    \big(1 \ - \ \eta \cdot a \cdot \E_{\mathcal{D}}\big[\x[j]^2\big]\big) \ \cdot \ \W_1[i, j]
    \quad & \textrm{in the case of GD.} \\[0.2cm]
    \big(
    1 \ - \ \eta \cdot a \cdot \E_{\mathcal{D}}\big[\x[j]^2\big] 
    \ - \ \eta \cdot \tfrac{\eta}{2b} \cdot a^2 \cdot c^2
    \big) \ \cdot \ \W_1[i, j]
    \ \quad & \textrm{in the case of SGD.}
\end{cases}
\end{split}
\end{equation*}

\end{theorem}
The meaning of this theorem is that the size of the weights of the first layer interacting with irrelevant components is non-increasing along the trajectory, and often decreasing.
Moreover, not only they are shrunk, in expectation they are shrunk more by SGD without replacement.
Precisely, in the notations of Theorem \ref{theo:shrinking} above, $c$ is a quantity 
related to the fourth moment of the irrelevant features over the dataset. The value $a$, instead, is a positively weighted sum of the singular values of the subsequent layers. 
More precisely, if we denote by $\tilde \W := \W_L\cdot \W_{L-1} \cdot \ldots \cdot \W_2$, then $a$ is the corresponding singular value of $\tilde \W$ or analogously $a:= \tilde \W^\top \tilde \W [i,i]$.
For more details, see Propositions \ref{prop:1} and \ref{prop:2}. 

The speed up of mini-batch SGD is an effect of its second-order implicit regularization implied by the fact that $\E_B[\x\x^\top]$ on every sampled batch $B$ is generally different from $\E_\mathcal{D}[\x\x^\top]$. 
We also find that this regularizing term is linearly bigger for higher values of $\eta/ b$, see also Figures \ref{fig:abstract:support:sine:gw}, \ref{fig:text:support:linear:irelnorms}, \ref{fig:text:vision:mnist:hist}, and \ref{fig:implicit:reg:dynamics}. This dependency on $\eta$ and $b$ matches the one of empirical observations in the literature \cite{goyal_accurate_2017,jastrzkebski2017three,he_control_2019}.

\section{Convergence: only with mini-batch SGD}
\label{section:SGD_better}

It is not surprising that the function represented by the neural network depends on the same input feature as the target function. What is surprising is that those relevant input features are identified in the first layer.
We show here that this is not the case for full-batch GD. SGD, however, does provably better than GD. Precisely, SGD always achieves smaller sizes for the weights in the first layer associated with irrelevant components, and anneals them when it trains in an oscillatory regime.

\subsection{GD does not learn the support in the first layer}
In the setting of Theorem \ref{theo:shrinking}, assume $a = \tilde \W^\top \tilde \W[i,i] = 0$ (in the following layers the output of the neuron $i$ is zeroed out) before the weight $\W_1[i,j]$ converges to zero. This implies that the subsequent layers stopped looking at that neuron, thus 1) the training (and the shrinking) stops on $\W_1[i,j]$ which is constant for the rest of the training, and 2) the subsequent layers are now responsible for identifying the support instead of the first one. In this case, indeed, the weights of the first layer do not give information about the support of the learned function.
We show here that this is very often the case for full-batch GD whereas it is not for mini-batch SGD.
In particular, we prove in the following result that full-batch GD rarely converges to a lower-rank first weight matrix. Indeed, the shrinking effects usually loses power exponentially fast. An example is the easy case of a diagonal networks initialized with any of the used initialization scheme, or more generally when every parameter is sampled independently from a distribution.

\begin{proposition}
\label{prop:GD}
Let $f_\theta$ be an either linear or ReLU diagonal network. Assume each weight is randomly initialized.
When trained with gradient descent or gradient flow the neural network learns the support in $O(\eta^{-1})$ time, but the probability that it identifies it in the first layer is smaller than $\mathrm{depth}^{-(d-r)}$.
\end{proposition}

\subsection{SGD does it because of its implicit bias}
Mini-batch SGD\footnote{Proposition \ref{prop:SGD} holds either for SGD with and without replacement. SGD without replacement converges marginally faster thanks to the direct application of the work in \cite{beneventano_trajectories_2023}. However, we did not quantify it.}, contrary to full-batch GD, does identify the support in the first layer even when either the training is unstable, noise is present on the labels, or the model is misspecified.

\begin{proposition}[Corollary of Theorem \ref{theo:time}]
\label{prop:SGD}
Assume that the $j$th component is irrelevant $\E_{\mathcal{D}}\big[y[j] \x[j]\big]=0$ but misspecified $\mathcal{R}_j^2 := \E\big[y[j]^2x[j]^2\big] > 0$ over our dataset. Let $f_\theta$ be a diagonal linear network. Assume the step size $0 < \eta < \eta_{\max}$\footnote{$\eta$ is smaller than 2 divided by the largest eigenvalue of the Hessian at initialization. See Appendix \ref{section:SGD}.}.
When trained with mini-batch SGD the neural network learns the support in the first layer with high probability in $t$ steps. Where $t = \tilde O \left( \frac{b}{\eta^2 \mathcal{R}_j} \right)^{2^{\mathrm{depth}-1}}$.
\end{proposition}
This is a direct corollary of Theorem \ref{theo:time} in Section \ref{section:implicit_reg}.
The key technical ingredient of the proof is checking that some quantities shrink as a geometric (multinomial) random walk and applying an Hoeffding inequality on the trajectories.
Moreover, we can add that in the case of two layers the time of convergece of the irrelevant weights of the first layer is $O(b/\eta^2 \mathcal{R}_j^2)$ where the $O$ hides logarithmic dependencies on initialization and the constants of the Hoeffding inequality.


\section{The reason is implicit SGD regularization}
\label{section:implicit_reg}
Neural networks learn the support in two different phases. Initially, they learn the representation that is the closest to the initialization. Secondly, the algorithm implicitly regularizes, leading to the identification of the support more prominently in the first layer, see Fig.\ \ref{fig:text:support:linear:irelnorms}. In this section we characterize these phases and we explore the underlying implicit regularization mechanisms of SGD.

\subsection{The 2 phases of the dynamics}
\paragraph{Phase 1: Learning the target function.}
The dynamics here is dominated by the minimization of the loss. This is the phase in which in Theorem \ref{theo:shrinking} we have $a > 0$. Neural networks, at the beginning, learn the support of the target function because they are learning the target function. Initialization in this phase strongly biases the dynamics, indeed, if the weights in hidden layers are smaller, the neural network quickly learns to identify the support in those layers, see Proposition \ref{prop:GD}.
It is important to note that when weight decay is added the support in the first layer is also learned in this phase.

\paragraph{Phase transition.}
For full-batch GD, this initial phase is the whole training. In contrast, after $O(1/\eta)$ steps mini-batch SGD often ceases to decrease the loss and begins to oscillate around the manifold of minima.
As highlighted by the literature in convex optimization \cite{mishchenko_random_2020}, this occurs at a distance of $O(\eta)$ from the manifold for SGD with replacement and $O(\eta^2)$ for SGD without replacement. SGD is now selecting a precise minimum on the manifold rather than converging further towards the manifold. This marks the beginning of the second phase for the training dynamics.

\paragraph{Phase 2: The first layer aligns with the support.}
While for training time smaller than $O(\eta^{-1})$, in the first phase of the dynamics neural networks are learning the target function through optimizing the loss, in this second phase what matters is how the algorithm travels through flat areas, not the better or worse convergence rates.
This phase is powered by the way the single algorithm optimizes the loss. Not by the fact that the loss is optimized.
It is thus inherently about the way these algorithms descend your landscape, not about the fact that they are descending the landscape. Thus this phase of alignment happens only in the case in which either regularization is explicitly injected or the algorithm implicitly regularizes by oscillating in a certain way or favoring certain directions.
This second phase of the training thus does not exists for gradient flow or GD, unless unstable \cite{damian_self-stabilization_2023} but exists and it is well documented for SGD without replacement \cite{beneventano_trajectories_2023}, SGD with replacement, and label noise \cite{damian_label_2021,li_what_2022,shalova_singular-limit_2024}.
After the target function is learned, in this phase, the algorithm picks a different set of parameters that represent that function. Precisely, the algorithm moves along the manifold of minima towards one that represents it better relative to the implicit bias (e.g., minimum norm or more degenerate Hessian). Identifying the support in the first layer is just an implication of this.
We explain what happens in the following Theorem \ref{theo:time}.
Denote the weights of the model interacting with $j$th component are
$\W_L[j,j], \W_{L-1}[j,j], \ldots ,\W_2[j,j], \W_1[j,j]$ for a diagonal linear network.
\begin{theorem}
\label{theo:time}
    Assume that the $j$th component is irrelevant $\E_{\mathcal{D}}\big[y[j] \x[j]\big]=0$. Assume that $\min_\theta \E_{(\x,y)\in \mathcal{D}}\big[ f_\theta(\x)[j]\x[j] - y[j]\x[j] \big] = c > 0$. Let $f_\theta$ be a diagonal linear network. Let $0<\eta<\eta_{\max}$.
    The weight $\W_*[j,j]$ which is the $i$th smallest in absolute value at initialization is zeroed out in time that is double exponential in $i$, in high probability over the SGD dynamics and any used random initialization scheme. More precisely
    Let $\delta_1, \delta_2 > 0$. Let $\alpha = \|\W_*[j,j]\|$ at initialization. Then with probability higher than $1-\delta_2$, we have that $\|\W_*[j,j]\| \leq \delta_1$ after a number of steps that is 
    \vspace{-0.3cm}
    \[
    O\left( 
    \eta^{2^{i-1}}c^{2^{i-1}} \big( \log(\alpha) - \log(\delta_2) - \log(\delta_1)\big)
    \right)
    \]
    for either SGD with and without replacement. However strictly faster for SGD without replacement.
\end{theorem}

\subsection{What powers the implicit regularization effect}
The second phase is characterized by the presence of oscillations and instabilities in the trajectory. What causes these instabilities can be either: (i) noise in the label or misspecified data, (ii) large variance in the inputs, or (iii) the fact that the learning rate is too large. In practice, the effect of (iii) the size of the step size, is to only stress the oscillations due to the inputs (i) or the outputs (ii).

\begin{figure}[ht]
\begin{minipage}{.33\textwidth} 
\centering
\includegraphics[height = 1\linewidth]{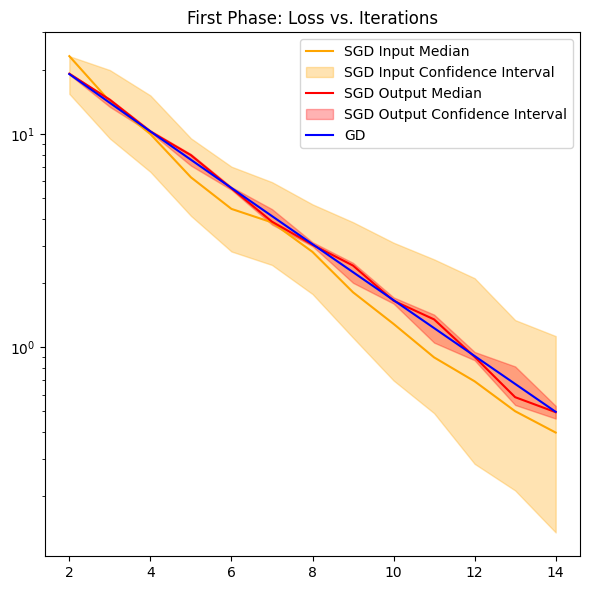}
\end{minipage}
\hfill
\begin{minipage}{.67\textwidth} 
\centering
\includegraphics[height = 0.5\linewidth]{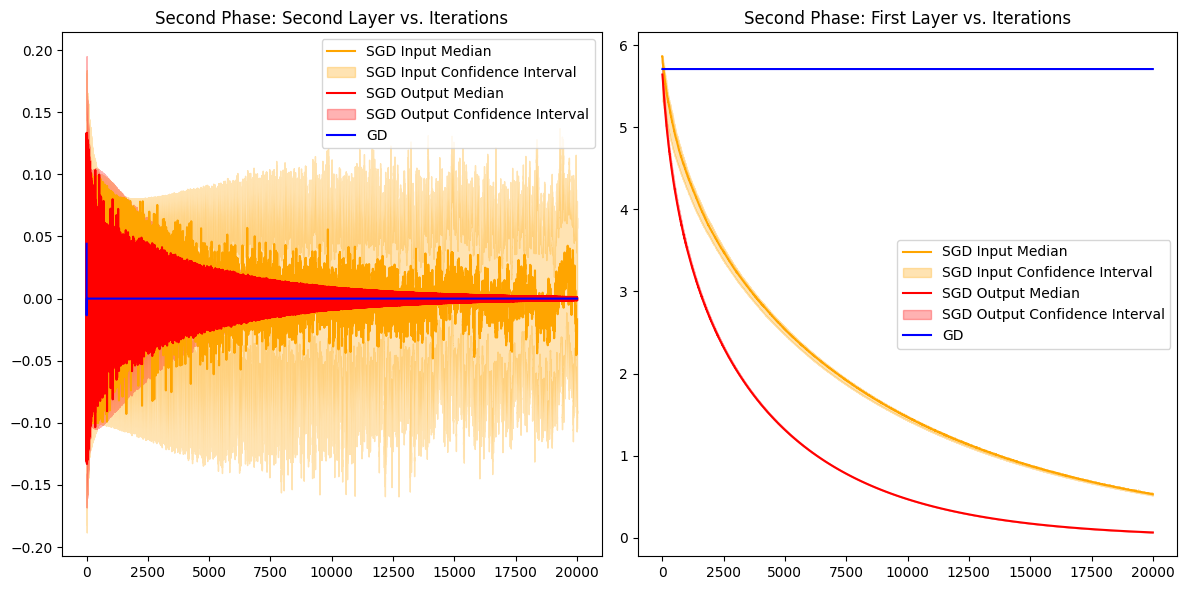}
\end{minipage}
\caption{
\textbf{Convergence of $a,b$ in Illustrative Example of Implicit SGD Regularization.}
}
\label{fig:implicit:reg:dynamics}
\end{figure}

\paragraph{(i) Oscillating on the inputs.}
In Theorem \ref{theo:shrinking} we find that SGD without replacement speeds up the convergence. 
This happens because the covariance matrices on the batches averages to the covariance matrix on the whole dataset $\mathcal{D}$. These covariance matrices define the speed of convergence of the weights. At every step the iteration of GD shrinks the weights by multiplying them by 1 minus a positive constant $\alpha$. In the case of SGD without replacement we have oscillation around alpha that cancel out implying an effect as the following
\begin{equation}
\label{eq:SGD_speed}
\underbrace{( 1 - \alpha + \delta )(1- \alpha +\delta)}_{\substack{\text{Shrinking due to SGD}}}
\quad = \quad
( 1 - \alpha )^2 - \delta^2
\quad < \quad 
\underbrace{( 1 - \alpha )^2}_{\substack{\text{Shrinking due to GD}}}.
\end{equation}
This implies that SGD has an effective higher speed of convergence than GD, see the plot on the left of Fig.\ \ref{fig:implicit:reg:dynamics}. 
This effect respects the so called linear scaling rule \cite{goyal_accurate_2017,jastrzkebski2017three,he_control_2019}. Indeed, $\delta$ is linearly proportional to the learning rate squared divided by the batch size. Precisely, $\delta$ is the learning rate squared times how much the covariance matrix on a batch is different from the full-batch one. Thus proportional to the size of variance of the mini-batch covariace matrix (fourth moment of the inputs) divided by the batch size.

\paragraph{(ii) Oscillating on the outputs.}
This second implicit regularization effect of SGD does not impact the speed of convergence itself, but impacts the location of convergence.
It implies that the models converge to flatter minima and it is powered by either mispecification or presence of noise in the labels. 
Precisely, if the labels are noisy or misspecified, after a first phase of convergence, SGD will enter an oscillatory regime. The fact of being in this oscillatory regime implies that SGD travels the manifold.
Essentially, in Eq.\ \ref{eq:SGD_speed} above, this effect virtually is equivalent to increasing the size of $\alpha$. This is key because it does not only mean the convergence is faster, it means that convergence happens. Indeed, often $\alpha$ converges to zero very fast in the case of GD undermining the convergence of the weights of the first layer. However, with noisy or mispecified labels, it does not.

\section{Practical Implications}
\label{section:implications}

\subsection{SGD may not need a smart initialization}
Working with any gradient-based machine learning algorithm involves the tedious task of tuning the optimizer’s hyperparameters, such as the initialization of its parameters. Theorem \ref{theo:time}, Proposition \ref{prop:GD}, and Proposition \ref{prop:need4wd} suggest that SGD is robust to initialization, as GD with regularization, but unlike unregularized GD. Precisely, no matter the initialization the first weight converges to the true support of target function. To give more context, in the literature it has been studied for long how to initialize such that GD goes to good minima and how very particular balanced initializations bring to the lowest norm solution \cite{vaskevicius_implicit_2019,woodworth_kernel_2020,gidel_implicit_2019,arora_implicit_2019,arora_convergence_2019}. However, neural networks at initialization are balanced with probability $0$ and convergence for balanced networks may be extremely slow. 
Theorem \ref{theo:time}, instead, shows that the second order effect of mini-batch SGD, with a speed proportional to the ratio $\eta / \text{batch size}$, implies convergence toward the minima to which \textit{balanced initialization} converges, no matter the initialization.
\textit{Our result is inherently a result of uncorrelation between initialization and trained model}.
This is also what we observe in the experiments, see Fig. \ref{fig:abstract:support:sine:gw} and \ref{fig:text:support:linear:irelnorms} where shrinking the batch size we also shrink the dependence from the initialization. See also Appendix \ref{appendix:support:synth} for more experiments where, ablating on the initialization, we observe that the minimum found from GD depends a lot on initialization, while the one found that the dependence on initialization after a certain time of training is increasing in the batch size.

\subsection{Enhanced features interpretability}
\begin{figure}[ht]
\centering
\includegraphics[width=\textwidth]{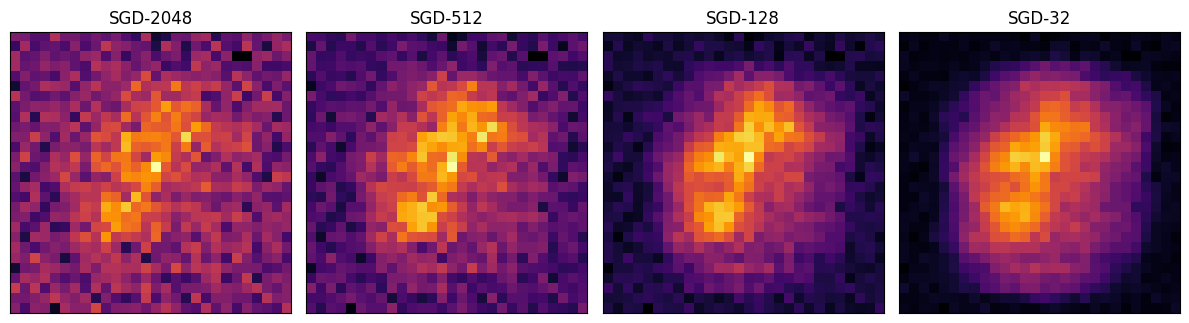}
\caption{\textbf{Where MLP model looks at on MNIST dataset.}}
\label{fig:vision:mnist:focusmap}
\end{figure}

While deep networks are remarkably effective in features extraction and regression using their final hidden representations, yet interpreting the specific features these models rely on during inference time remains challenging. We argue that that the bias in mini-batch SGD towards learning the support in the first layer may offer valuable insights and form a basis for understanding the features networks use. Figure \ref{fig:vision:mnist:focusmap} shows the areas within average MNIST digit inputs that a trained MLP model attends to depending on the batch size. Although each of these optimizers achieve comparable test losses, smaller batch sizes enhance the network's learning of support in the first layer, thereby improving not only robustness but interpretability. 


\section{Extensions: The case of ReLU \& the benign effect of weight decay}
\label{section:ReLU}
\paragraph{Extension to nonlinear activation.}
In the case of networks with nonlinear activation, the definition of irrelevant input features is more nuanced. Imagine, the example of $y(\x)=\x^2 \in \R^d$. If the the $\x$ are balanced around zero, $\E[y(\x)\x] = \E[\x^3] = 0$ but $\E[\x^2]>0$, thus this function is learnable by ReLU networks but it is mispecified for linear networks where the component is treated as an irrelevant one.
In particular, various theorems of universality of approximations imply that generally functions are not misspecified by neural networks. The irrelevant component $\x[j]$ should thus be defined as those that are irrelevant \textit{for all possible values attained} by the other (relevant) components of $\x$, unlike for linear networks, where we were just \textit{averaging} on the other components.
\begin{equation}
\label{eq:irrelevant_ReLU}
\E_{\mathcal{D}}\left[
y(\x) \Big| \substack{\text{relevant}\\\text{components}} \ \x[:r]
\right] \quad = \quad 0
\qquad \text{instead of} \qquad 
\E_{\mathcal{D}}\left[
y(\x)\x[j] \right] \quad = \quad 0
\end{equation}
Learning the support in the first layer, therefore, means that the weights of the first layer have to converge to 0 if interacting with either the components (i) in the kernel of $\Lambda_i$ that are not spanned by the data, and (ii) for which the balance assumption above in Eq.\ \ref{eq:irrelevant_ReLU} holds. This is provably the case when we add weight decay but it is not without any regularization.
\begin{proposition}
\label{prop:need4wd}
Let $f_\theta$ be a neural network with either linear or ReLU activation $\sigma$. Assume $0 < \eta < \eta_{max}$\footnote{We properly define the instability threshold in Proposition \ref{prop:3}. That said, it is bigger than the biggest eigenvalue of the Hessian of the loss, minus the size of the weight decay $\lambda > 0$.}. Then GD or SGD with weight decay of intensity $\lambda>0$ learn the support in the first layer with exponential speed, without weight decay they do not.
\end{proposition}

\paragraph{The case of ReLU.}
For the sake of the explanation, let us work with a shallow ReLU network. For every input $x$ let us denote by $D^x$ the diagonal matrix diagonal entry $1$ if the corresponding neuron in the layer $i$ is on, $0$ if it is off. The effect of an SGD iteration on the batch $\mathcal{B}$ on the weights $\W_1[:, r+1:]$ corresponding to the irrelevant components can be rewritten as
\begin{equation*}
\begin{split}
    - \ 
    \frac{\eta}{n} \sum_{(\x,y) \in \mathcal{B}}
    \underbrace{(D^\x \W_2^\top \W_2 D^\x)}_{PSD} \cdot \W_1[:,r+1:] \cdot \underbrace{\x[r+1:]\x[r+1:]^\top}_{PSD} 
    \quad &\leftarrow \quad \substack{\text{non-increasing on the}\\ \text{irrelevant weights}}
    \\- \ 
    \frac{\eta}{n} \sum_{(\x,y) \in \mathcal{B}}
    \underbrace{D^\x \W_2^\top}_{\text{a matrix}} \cdot \underbrace{\big( \W_2 D^\x \W_1[:,:r] \cdot \x[:r] - y\big)}_{\text{residuals on the relevant input features}}
    \cdot \underbrace{\x[r+1:]^\top}_{\text{irrelevant}}
    \quad \ \ \ &\leftarrow \quad \substack{\substack{\text{ReLU} \implies \text{unpredictable}\\ \text{Linear} \implies \text{no effect }\phantom{aaa}}}
\end{split}
\end{equation*}
In the linear case we could conclude that the second line is $0$ in expectation over the data or that it is centered in zero and induces benign oscillations for SGD. In the case of ReLU we can not as $\W_2 D^\x$, unlike $\W_2$, is not the same on every data point. This implies that there are cases in which some entries of $\W_1[:,r+1:]$ increase along the training without weight decay. More precisely, the dynamics on the rows of $\W_1$ will be as follows depending on what data points activate the corresponding neuron. On the $r$ relevant components it learns. On the $u$ components not spanned the updates above zero out, so the dynamics is stationary unless weight decay is present. On the $i$ irrelevant components that are spanned by the data either (i) the dynamics behaves as the linear case if the data points of $\mathcal{D}$ on which the neuron is active have small $(f_\theta(\x)-y)\cdot \x[r+1:]$, this increasingly happens when the target function is well approximated and when the condition in Eq.\ \ref{eq:irrelevant_ReLU} holds. In this case the weights relative to the irrelevant components are non-increasing because of the effect of the first line above, and decay strongly if weight decay is present; or (ii) the size increases or decreases substantially otherwise. 
For more about the case of ReLU see Appendix \S \ref{section:ReLU_app} and \ref{section:ReLU_failure} or the experiments in Appendix \ref{appendix:support:synth} where we show that some rows of the weights of the first layer effectively converge as for linear networks, others (the ones on which the irrelevant components are unbalanced) do not.


\section{Conclusion}
We explored how different optimization algorithms affect support selection, focusing on mini-batch SGD and GD. Our findings reveal that mini-batch SGD and regularized GD promote the learning of relevant input features in the first layer, unlike unregularized GD. This property implies lower dependency on initialization and enhanced interpretability for neural networks. Importantly, our results serve as a foundational block for understanding how real-world neural networks represent features, offering insights crucial for both theoretical exploration and practical application. These insights deepen our understanding of learning dynamics in neural networks and guide the selection of optimization strategies to achieve more effective and interpretable models.

\paragraph{Broader Impact.}
Our findings enhance the understanding of how features are learned by neural networks in practice, potentially aiding the development of a theory of neural networks. This theoretical foundation is key to certifying neural networks for high-stakes real-world applications. Although our work is theoretical and focuses on understanding optimization effects, it underscores the importance of training strategies and hyperparameter tuning in effectively capturing the underlying structure of data. However, our work focuses on understanding these optimization effects, without immediate direct societal impacts.

\paragraph{Limitations.}
Despite the weak assumptions on the data, the main limitations of our theory concern the architectures considered. Specifically, we focused on linear neural networks for Theorem \ref{theo:shrinking}, diagonal networks for Proposition \ref{prop:GD}, and diagonal linear networks for Proposition \ref{prop:SGD}. To address these limitations, we discussed extensions to ReLU networks and identified the associated challenges in Section \ref{section:ReLU}. Furthermore, extensive empirical experiments demonstrate that our findings hold across a wide range of MLPs with nonlinear activation functions.


\begin{ack}
We thank Boris Hanin, Jason Lee, Alex Damian, Tomer Galanti, Valeria Ambrosio, and Arseniy Andreyev for many relevant discussions and comments. This material is based on the work supported by the Center for Minds, Brains and Machines (CBMM) at MIT funded by NSF STC award CCF-1231216. Andrea Pinto was also funded by Fulbright Scholarship to pursue this research.
\end{ack}


\clearpage

\bibliography{reference}

\begin{thebibliography}{10}

\bibitem{arora_convergence_2019}
Sanjeev Arora, Nadav Cohen, Noah Golowich, and Wei Hu.
\newblock A {Convergence} {Analysis} of {Gradient} {Descent} for {Deep} {Linear} {Neural} {Networks}, October 2019.
\newblock arXiv:1810.02281 [cs, stat].

\bibitem{arora_implicit_2019}
Sanjeev Arora, Nadav Cohen, Wei Hu, and Yuping Luo.
\newblock Implicit regularization in deep matrix factorization.
\newblock In {\em Advances in {Neural} {Information} {Processing} {Systems}}, pages 7411--7422, 2019.

\bibitem{lecun1998mnist}
Yann LeCun, L{\'e}on Bottou, Yoshua Bengio, and Patrick Haffner.
\newblock Gradient-based learning applied to document recognition.
\newblock {\em Proceedings of the IEEE}, 86(11):2278--2324, 1998.

\bibitem{he2015deep}
Kaiming He, Xiangyu Zhang, Shaoqing Ren, and Jian Sun.
\newblock Deep residual learning for image recognition.
\newblock {\em Proceedings of the IEEE conference on computer vision and pattern recognition}, pages 770--778, 2015.

\bibitem{krizhevsky2009learning}
Alex Krizhevsky.
\newblock Learning multiple layers of features from tiny images.
\newblock Technical report, University of Toronto, 2009.

\bibitem{poggiofraser2024}
Tomaso Poggio and Maia Fraser.
\newblock Compositional sparsity of learnable functions.
\newblock {\em Bulletin of the American Mathematical Society}, 2024.

\bibitem{10.1214/19-AOS1875}
Johannes Schmidt-Hieber.
\newblock {Nonparametric regression using deep neural networks with ReLU activation function}.
\newblock {\em The Annals of Statistics}, 48(4):1875 -- 1897, 2020.

\bibitem{kohler2020rate}
Michael Kohler and Sophie Langer.
\newblock On the rate of convergence of fully connected very deep neural network regression estimates, 2020.

\bibitem{Memo144}
Mengjia Xu, Tomer Galanti, Akshay Rangamani, Andrea Pinto, and Tomaso Poggio.
\newblock The janus effects of sgd vs gd: high noise and low rank.
\newblock Technical report, Center for Brains, Minds and Machines (CBMM), 2023.

\bibitem{abbe_provable_2023}
Emmanuel Abbe, Elisabetta Cornacchia, and Aryo Lotfi.
\newblock Provable {Advantage} of {Curriculum} {Learning} on {Parity} {Targets} with {Mixed} {Inputs}, June 2023.
\newblock arXiv:2306.16921 [cs, stat].

\bibitem{cornacchia_mathematical_2024}
Elisabetta Cornacchia and Elchanan Mossel.
\newblock A {Mathematical} {Model} for {Curriculum} {Learning} for {Parities}, April 2024.
\newblock arXiv:2301.13833 [cs].

\bibitem{abbe_provable_2019}
Emmanuel Abbe and Colin Sandon.
\newblock Provable limitations of deep learning, April 2019.
\newblock arXiv:1812.06369 [cs, math, stat].

\bibitem{abbe_poly-time_2020}
Emmanuel Abbe and Colin Sandon.
\newblock Poly-time universality and limitations of deep learning, January 2020.
\newblock arXiv:2001.02992 [cs, math, stat].

\bibitem{abbe_merged-staircase_2022}
Emmanuel Abbe, Enric Boix-Adsera, and Theodor Misiakiewicz.
\newblock The merged-staircase property: a necessary and nearly sufficient condition for {SGD} learning of sparse functions on two-layer neural networks, February 2022.
\newblock arXiv:2202.08658 [cs, stat].

\bibitem{abbe_sgd_2023}
Emmanuel Abbe, Enric~Boix Adserà, and Theodor Misiakiewicz.
\newblock {SGD} learning on neural networks: leap complexity and saddle-to-saddle dynamics.
\newblock In {\em Proceedings of {Thirty} {Sixth} {Conference} on {Learning} {Theory}}, pages 2552--2623. PMLR, July 2023.
\newblock ISSN: 2640-3498.

\bibitem{damian_neural_2022}
Alexandru Damian, Jason Lee, and Mahdi Soltanolkotabi.
\newblock Neural {Networks} can {Learn} {Representations} with {Gradient} {Descent}.
\newblock In {\em Proceedings of {Thirty} {Fifth} {Conference} on {Learning} {Theory}}, pages 5413--5452. PMLR, June 2022.
\newblock ISSN: 2640-3498.

\bibitem{damian_optimal_2023}
Alex Damian, Eshaan Nichani, Rong Ge, and Jason~D Lee.
\newblock Optimal {Sample} {Complexity} for {Learning} {Single} {Index} {Models}.
\newblock May 2023.

\bibitem{bietti_learning_2022}
Alberto Bietti, Joan Bruna, Clayton Sanford, and Min~Jae Song.
\newblock Learning {Single}-{Index} {Models} with {Shallow} {Neural} {Networks}.
\newblock October 2022.

\bibitem{bottou_stochastic_2012}
Léon Bottou.
\newblock Stochastic {Gradient} {Descent} {Tricks}.
\newblock In Grégoire Montavon, Geneviève~B. Orr, and Klaus-Robert Müller, editors, {\em Neural {Networks}: {Tricks} of the {Trade}: {Second} {Edition}}, Lecture {Notes} in {Computer} {Science}, pages 421--436. Springer, Berlin, Heidelberg, 2012.

\bibitem{lecun_efficient_2012}
Yann~A LeCun, Léon Bottou, Genevieve~B Orr, and Klaus-Robert Müller.
\newblock Efficient backprop.
\newblock In {\em Neural networks: {Tricks} of the trade}, pages 9--48. Springer, 2012.

\bibitem{bengio_practical_2012}
Yoshua Bengio.
\newblock Practical recommendations for gradient-based training of deep architectures, September 2012.
\newblock arXiv:1206.5533 [cs].

\bibitem{liu_toward_2020}
Chaoyue Liu, Libin Zhu, and Mikhail Belkin.
\newblock Toward a theory of optimization for over-parameterized systems of non-linear equations: the lessons of deep learning.
\newblock {\em arXiv:2003.00307 [cs, math, stat]}, February 2020.
\newblock arXiv: 2003.00307.

\bibitem{nguyen_connected_2019}
Quynh Nguyen.
\newblock On {Connected} {Sublevel} {Sets} in {Deep} {Learning}.
\newblock {\em arXiv:1901.07417 [cs, stat]}, May 2019.
\newblock arXiv: 1901.07417.

\bibitem{zhang_embedding_2022}
Yaoyu Zhang, Zhongwang Zhang, Tao Luo, and Zhi-Qin~John Xu.
\newblock Embedding {Principle} of {Loss} {Landscape} of {Deep} {Neural} {Networks}, January 2022.
\newblock arXiv:2105.14573 [cs, stat].

\bibitem{neyshabur_search_2015}
Behnam Neyshabur, Ryota Tomioka, and Nathan Srebro.
\newblock In {Search} of the {Real} {Inductive} {Bias}: {On} the {Role} of {Implicit} {Regularization} in {Deep} {Learning}.
\newblock {\em arXiv:1412.6614 [cs, stat]}, April 2015.

\bibitem{keskar_large-batch_2016}
Nitish~Shirish Keskar, Dheevatsa Mudigere, Jorge Nocedal, Mikhail Smelyanskiy, and Ping Tak~Peter Tang.
\newblock On large-batch training for deep learning: {Generalization} gap and sharp minima.
\newblock {\em arXiv preprint arXiv:1609.04836}, 2016.

\bibitem{jastrzebski_catastrophic_2021}
Stanisław Jastrzębski, Devansh Arpit, Oliver Astrand, Giancarlo Kerg, Huan Wang, Caiming Xiong, Richard Socher, Kyunghyun Cho, and Krzysztof Geras.
\newblock Catastrophic {Fisher} {Explosion}: {Early} {Phase} {Fisher} {Matrix} {Impacts} {Generalization}.
\newblock {\em arXiv:2012.14193 [cs, stat]}, June 2021.
\newblock arXiv: 2012.14193.

\bibitem{beneventano_trajectories_2023}
Pierfrancesco Beneventano.
\newblock On the {Trajectories} of {SGD} {Without} {Replacement}, December 2023.
\newblock arXiv:2312.16143 [cs, math, stat].

\bibitem{goyal_accurate_2017}
Priya Goyal, Piotr Dollár, Ross Girshick, Pieter Noordhuis, Lukasz Wesolowski, Aapo Kyrola, Andrew Tulloch, Yangqing Jia, and Kaiming He.
\newblock Accurate, large minibatch sgd: {Training} imagenet in 1 hour.
\newblock {\em arXiv preprint arXiv:1706.02677}, 2017.

\bibitem{jastrzkebski2017three}
Stanis{\l}aw Jastrz{\k{e}}bski, Zachary Kenton, Devansh Arpit, Nicolas Ballas, Asja Fischer, Yoshua Bengio, and Amos Storkey.
\newblock Three factors influencing minima in {SGD}.
\newblock {\em arXiv preprint arXiv:1711.04623}, 2017.

\bibitem{he_control_2019}
Fengxiang He, Tongliang Liu, and Dacheng Tao.
\newblock Control {Batch} {Size} and {Learning} {Rate} to {Generalize} {Well}: {Theoretical} and {Empirical} {Evidence}.
\newblock In {\em Advances in {Neural} {Information} {Processing} {Systems}}, volume~32. Curran Associates, Inc., 2019.

\bibitem{gunasekar_implicit_2017}
Suriya Gunasekar, Blake~E Woodworth, Srinadh Bhojanapalli, Behnam Neyshabur, and Nati Srebro.
\newblock Implicit regularization in matrix factorization.
\newblock In {\em Advances in {Neural} {Information} {Processing} {Systems}}, pages 6151--6159, 2017.

\bibitem{neyshabur_exploring_2017}
Behnam Neyshabur, Srinadh Bhojanapalli, David Mcallester, and Nati Srebro.
\newblock Exploring {Generalization} in {Deep} {Learning}.
\newblock In I.~Guyon, U.~V. Luxburg, S.~Bengio, H.~Wallach, R.~Fergus, S.~Vishwanathan, and R.~Garnett, editors, {\em Advances in {Neural} {Information} {Processing} {Systems} 30}, pages 5947--5956. Curran Associates, Inc., 2017.

\bibitem{gidel_implicit_2019}
Gauthier Gidel, Francis Bach, and Simon Lacoste-Julien.
\newblock Implicit {Regularization} of {Discrete} {Gradient} {Dynamics} in {Linear} {Neural} {Networks}.
\newblock In {\em Advances in {Neural} {Information} {Processing} {Systems}}, volume~32. Curran Associates, Inc., 2019.

\bibitem{vaskevicius_implicit_2019}
Tomas Vaskevicius, Varun Kanade, and Patrick Rebeschini.
\newblock Implicit {Regularization} for {Optimal} {Sparse} {Recovery}.
\newblock In {\em Advances in {Neural} {Information} {Processing} {Systems}}, pages 2968--2979, 2019.

\bibitem{woodworth_kernel_2020}
Blake Woodworth, Suriya Gunasekar, Jason~D. Lee, Edward Moroshko, Pedro Savarese, Itay Golan, Daniel Soudry, and Nathan Srebro.
\newblock Kernel and {Rich} {Regimes} in {Overparametrized} {Models}.
\newblock In {\em Proceedings of {Thirty} {Third} {Conference} on {Learning} {Theory}}, pages 3635--3673. PMLR, July 2020.
\newblock ISSN: 2640-3498.

\bibitem{damian_label_2021}
Alex Damian, Tengyu Ma, and Jason Lee.
\newblock Label {Noise} {SGD} {Provably} {Prefers} {Flat} {Global} {Minimizers}.
\newblock {\em arXiv:2106.06530 [cs, math, stat]}, June 2021.

\bibitem{li_what_2022}
Zhiyuan Li, Tianhao Wang, and Sanjeev Arora.
\newblock What {Happens} after {SGD} {Reaches} {Zero} {Loss}? --{A} {Mathematical} {Framework}.
\newblock {\em arXiv:2110.06914 [cs, stat]}, February 2022.

\bibitem{chen_stochastic_2023}
Feng Chen, Daniel Kunin, Atsushi Yamamura, and Surya Ganguli.
\newblock Stochastic {Collapse}: {How} {Gradient} {Noise} {Attracts} {SGD} {Dynamics} {Towards} {Simpler} {Subnetworks}, June 2023.

\bibitem{orvieto_explicit_2023}
Antonio Orvieto, Anant Raj, Hans Kersting, and Francis Bach.
\newblock Explicit {Regularization} in {Overparametrized} {Models} via {Noise} {Injection}, January 2023.
\newblock arXiv:2206.04613 [cs, stat].

\bibitem{shalova_singular-limit_2024}
Anna Shalova, André Schlichting, and Mark Peletier.
\newblock Singular-limit analysis of gradient descent with noise injection, April 2024.

\bibitem{yaida_fluctuation-dissipation_2018}
Sho Yaida.
\newblock Fluctuation-dissipation relations for stochastic gradient descent.
\newblock {\em arXiv preprint arXiv:1810.00004}, 2018.

\bibitem{li_validity_2021}
Zhiyuan Li, Sadhika Malladi, and Sanjeev Arora.
\newblock On the {Validity} of {Modeling} {SGD} with {Stochastic} {Differential} {Equations} ({SDEs}).
\newblock {\em arXiv:2102.12470 [cs, stat]}, June 2021.

\bibitem{smith_origin_2021}
Samuel~L. Smith, Benoit Dherin, David G.~T. Barrett, and Soham De.
\newblock On the {Origin} of {Implicit} {Regularization} in {Stochastic} {Gradient} {Descent}.
\newblock {\em arXiv:2101.12176 [cs, stat]}, January 2021.

\bibitem{haochen_shape_2020}
Jeff~Z. HaoChen, Colin Wei, Jason~D. Lee, and Tengyu Ma.
\newblock Shape {Matters}: {Understanding} the {Implicit} {Bias} of the {Noise} {Covariance}.
\newblock {\em arXiv:2006.08680 [cs, stat]}, June 2020.

\bibitem{roberts_sgd_2021}
Daniel~A. Roberts.
\newblock {SGD} {Implicitly} {Regularizes} {Generalization} {Error}.
\newblock {\em arXiv:2104.04874 [cs, stat]}, April 2021.
\newblock arXiv: 2104.04874.

\bibitem{paszke2019pytorch}
Adam Paszke, Sam Gross, Francisco Massa, Adam Lerer, James Bradbury, Gregory Chanan, Trevor Killeen, Zeming Lin, Natalia Gimelshein, Luca Antiga, et~al.
\newblock Pytorch: An imperative style, high-performance deep learning library, 2019.

\bibitem{tensorflow2015-whitepaper}
Mart{\'i}n Abadi, Ashish Agarwal, Paul Barham, Eugene Brevdo, Zhifeng Chen, Craig Citro, Greg~S Corrado, Andy Davis, Jeffrey Dean, Matthieu Devin, et~al.
\newblock Tensorflow: Large-scale machine learning on heterogeneous systems, 2015.
\newblock Software available from tensorflow.org.

\bibitem{mishchenko_random_2020}
Konstantin Mishchenko, Ahmed Khaled, and Peter Richtarik.
\newblock Random {Reshuffling}: {Simple} {Analysis} with {Vast} {Improvements}.
\newblock In {\em Advances in {Neural} {Information} {Processing} {Systems}}, volume~33, pages 17309--17320. Curran Associates, Inc., 2020.

\bibitem{damian_self-stabilization_2023}
Alex Damian, Eshaan Nichani, and Jason~D. Lee.
\newblock Self-{Stabilization}: {The} {Implicit} {Bias} of {Gradient} {Descent} at the {Edge} of {Stability}, April 2023.
\newblock arXiv:2209.15594 [cs, math, stat].

\bibitem{he2015delving}
Kaiming He, Xiangyu Zhang, Shaoqing Ren, and Jian Sun.
\newblock Delving deep into rectifiers: Surpassing human-level performance on imagenet classification, 2015.

\bibitem{deng2009imagenet}
Jia Deng, Wei Dong, Richard Socher, Li-Jia Li, Kai Li, and Li~Fei-Fei.
\newblock Imagenet: A large-scale hierarchical image database.
\newblock In {\em 2009 IEEE conference on computer vision and pattern recognition}, pages 248--255. IEEE, 2009.

\end{thebibliography}
\bibliographystyle{unsrt}


\newpage
\appendix
\section{Vision Data is Inherently Low-Dimensional}
\label{section:low_dimensional}

\begin{figure}[ht]
\centering
\includegraphics[width=0.75\textwidth]{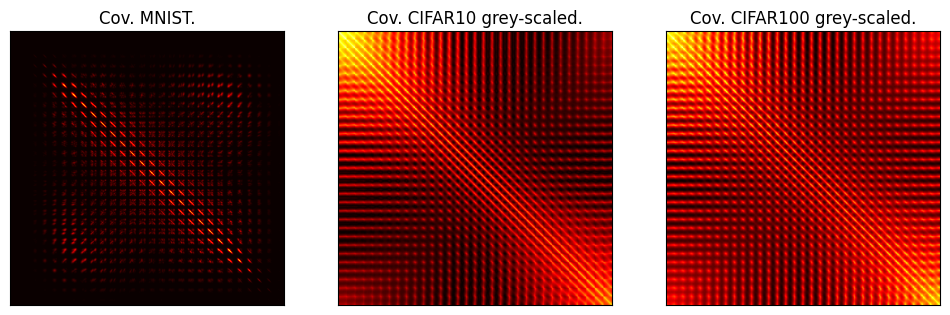}
\caption{Covariance matrices for the MNIST, CIFAR10, and CIFAR100 datasets in grayscale (average over RGB channels) representing the variance and correlation between pixel positions.}
\label{fig:lowdim:covariance}
\end{figure}

It is well-known in literature showing that the signal in the data is inherently low-dimensional or can be represented as such. In particular, here we show how although the input covariance for MNIST and CIFAR10 is high dimensional, as it has many sizeable singular values, most of the signal lies in a specific (non-linear) 9-dimensional space. This 9-dimensional non-linear space is representable by passing the input through a CNN. Indeed, most of the singular values of the inputs of the data, see Fig.\ \ref{fig:lowdim:covariance}, are of size $e+04$ or bigger. However, applying the non-linear function represented by 15 layers of convolutions and training an MLP on top we find that the MLP only depends on 9 input (input for the MLP) features of the data to classify with accuracy 92\%. This implies that there are 9 non-linear single-index models that describe the target function. Thus the target function has an inherently 9-dimensional support. Moreover, CNN with skipped connections represent all that matters and MLPs find the low dimensional representation.

\begin{figure}[ht]
\centering
\vspace{-5mm} 
\includegraphics[width=\textwidth]{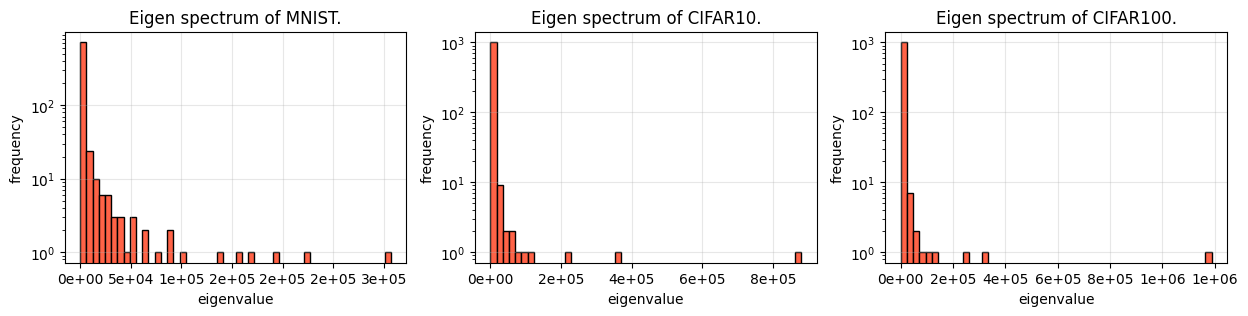}
\caption{Eigenvalue spectra of the MNIST, CIFAR10, and CIFAR100 datasets, depicting the frequency distribution of eigenvalues from covariance matrices shown in Fig.\ \ref{fig:lowdim:covariance}. MNIST shows 10 eigen spikes correlating with its 10 classes followed by a bulk concentrated towards zero, while CIFAR datasets exhibit same bulk punctuated by 3 eigen spikes reflecting the primary RGB channels.
}
\end{figure}

\begin{figure}[H]
\centering
\includegraphics[width=1.0\textwidth]{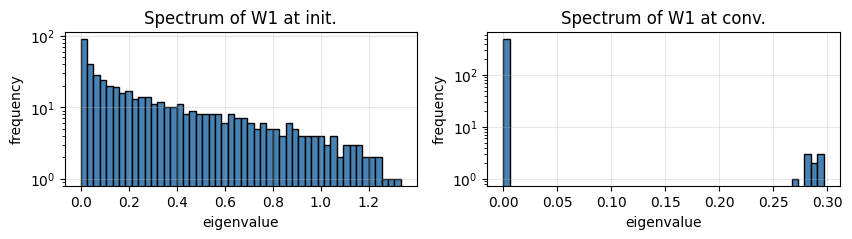}
\caption{Histogram of the eigenvalues of the first layer of a fine-tuned MLP head on top of a pretrained ResNet16 on CIFAR10. This figure shows that while CIFAR10 is very high-dimensional, the network only needs a 9-dimensional non-linear space to properly perform the classification.}
\label{fig:head:cifar10:ground:truth:histo}
\end{figure}

\newpage
\section{Implicit SGD Regularization : An Illustrative Example}
To show a heuristic of how it affects the training one could think of the model $f_{(a,b)}(x) = abx$. In this very easy example in the case in which we have two data points that have same $x$ and different $y$, so for instance $\mathcal{D}_1 = \{(1,-1), (1,1)\}$. Here, GD converges to the closest $a,b$ such that $a\cdot b=0$. However, SGD converges exactly to $a = b = 0$ which is the lowest norm solution. This can be proved by using the effect highlighted by \cite{beneventano_trajectories_2023} for SGD without replacement or by \cite{damian_label_2021} for SGD with replacement. The idea is that even if $a\cdot b = 0$ is converged, seeing the two misspecified data points in order induces the parameters to oscillate around the manifold of minima. This oscillation powers a second order effect that traverse this manifold in the direction of the lowest norm solution.

Imagine now to work with the dataset $\mathcal{D}_2 = \{(1,0), (3,0)\}$. Thus the model is well specified and there is no noise on the $y$, however we have noise on the $x$. Let us say $a > b$ WLOG at initialization. Then after one steps of GD $b \curvearrowleft (1 - 5\eta a^2) b$ and for SGD either $b \curvearrowleft (1 - 9\eta a^2) b$  or $b \curvearrowleft (1 - \eta a^2) b$. Iterating again and considering the two steps we have the following approximate rates of shrinking
\begin{equation}
\begin{cases}
(1 - 10\eta a^2\, \textbf{+\,25}\, \eta^2 a^4 + O(\eta^2b^2a^2) ) \qquad & \text{ for GD}.\\
(1 - 10\eta a^2\, \textbf{+\,9} \,\eta^2 a^4 + O(\eta^2b^2a^2) ) \qquad & \text{ for SGD}.
\end{cases}
\end{equation}

See the left plot of Fig.\ \ref{fig:implicit:reg:dynamics} for a plot of what is going on.

\begin{figure}[H]
\begin{minipage}{.5\textwidth}
\includegraphics[height=0.9\linewidth]{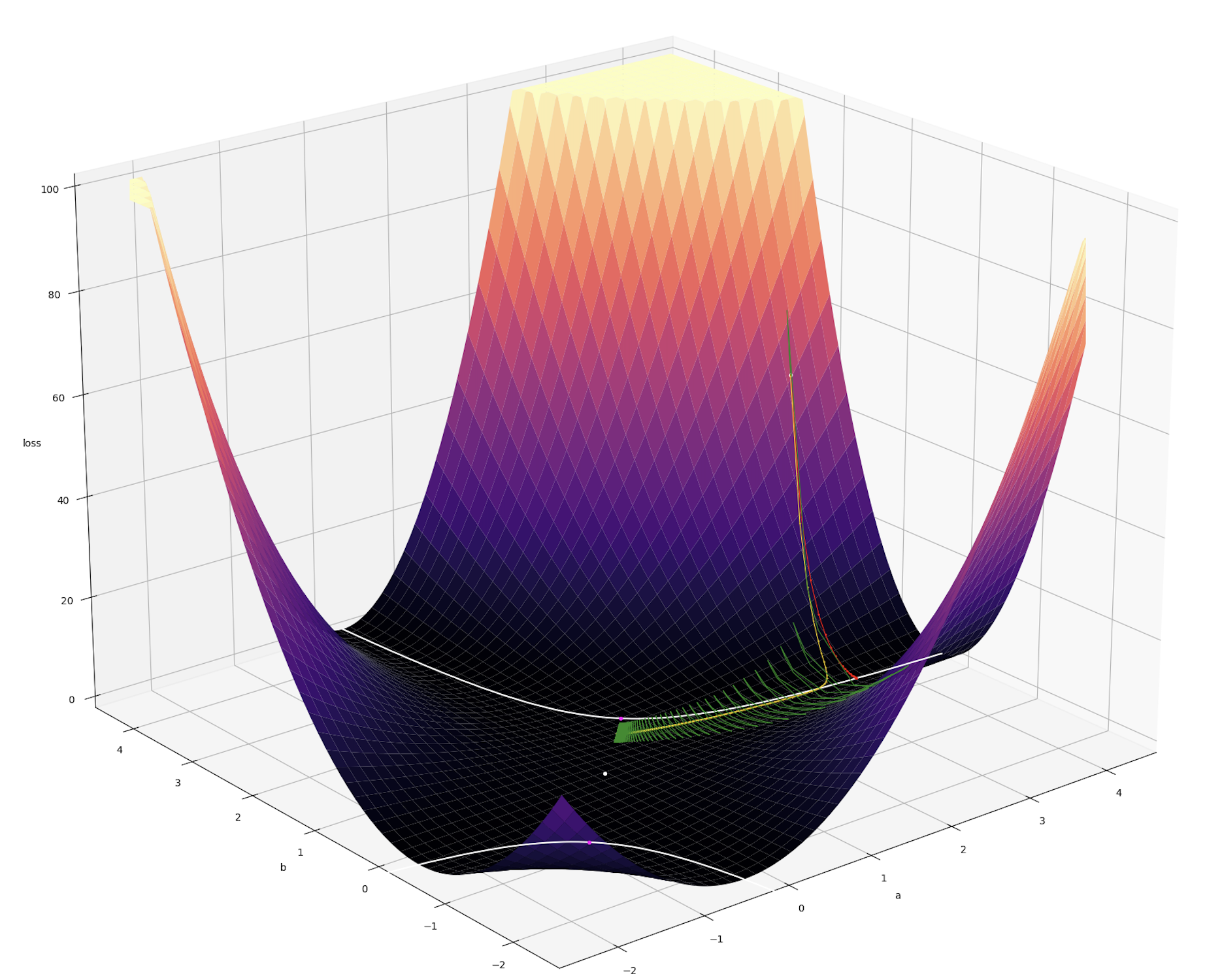}
\end{minipage}
\hfill
\begin{minipage}{.4\textwidth} 
\centering
\includegraphics[height=0.9\linewidth]{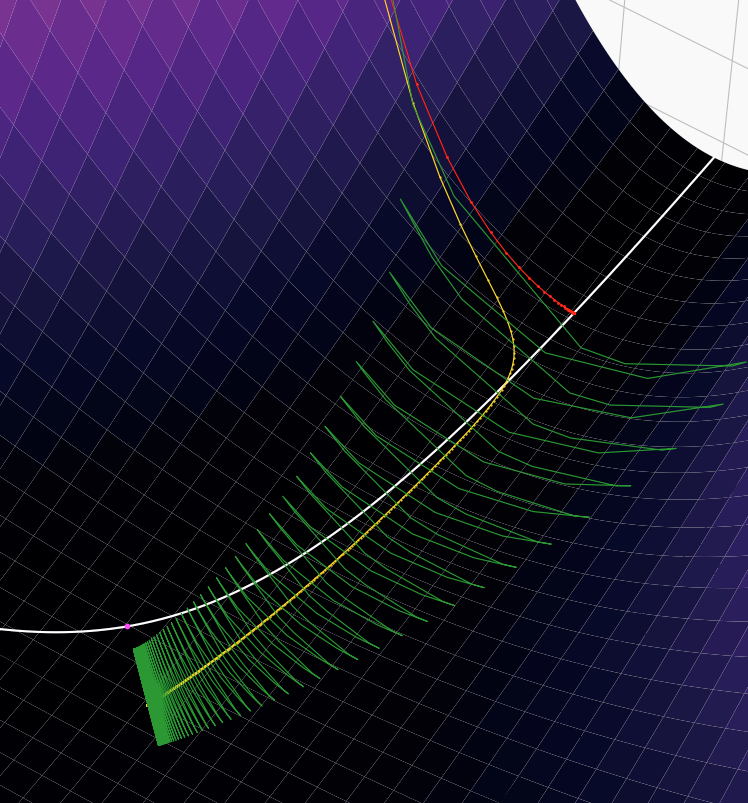}
\end{minipage}
\caption{\textbf{Visualization of Optimization Trajectories in a Toy Model.} We train $f(x) = abx$ using three distinct optimizers under MSE loss landscape: GD in red, SGD in green, and GD with weight decay in gold. The manifold of minima is indicated by the white line. We see that GD converges to the minima and ceases movement. In contrast, SGD exhibits second-order oscillations, leading to a persistent drift towards the minimum norm solution, and following regularized GD same direction.}
\label{fig:implicit:reg:landscape}
\end{figure}

\newpage
\section{Support Identification : Empirical Experiments}
\label{appendix:support:synth}
\paragraph{Picking up the support.} In this appendix, we detail the empirical experiments and ablation results supporting our primary claim. That is, deep neural networks (MLPs) perform target support selection in their first hidden layer when trained with SGD or with GD coupled with regularization. This property in mini-batch SGD is due to a second order regularization term shrinking the weights associated with irrelevant components of inputs to zero (See Theorem \ref{theo:shrinking}). While on the other hand, the same network (with same initialization) with vanilla GD achieves similar loss, but its first layer is not picking up the support of the target function. Instead, vanilla performs features engineering and support selection at the same time across all the layers of the network. Each GD layer has a given (equal) probability of eliminating irrelevant features of the input (See Proposition \ref{prop:GD}).

\paragraph{Risk as objective function.}
In the following subsection, we will be training MLPs of depth $d$ of the form $f_d(\x) = \W_d \sigma \ldots \W_2 \sigma \W_1 \x$ where the activation function $\phi(.)$ is either the identity or ReLU.
These networks are trained using a range of optimizers minimizing the (regularized) empirical risk.
\begin{align*}
\textrm{(ER)} \quad \hat{\mathcal{R}}_\mathcal{D}(f_\theta)~:=~\frac{1}{n}\sum_{i=1}^{n}\ell(f_\theta(x_i), y_i)
\quad \textrm{or} \quad
\textrm{(RER)} \quad \hat{\mathcal{R}}_\mathcal{D}^\lambda(f_\theta)~:=~\frac{1}{n}\sum_{i=1}^{n}\ell(f_\theta(x_i), y_i) + || \theta ||_F^2
\end{align*}
where $\lambda > 0$ is a predefined hyperparameter and $\|\cdot\|_F$ denotes the Frobenius norm. (ER) refers to the empirical risk and (RER) the regularized empirical risk of a parametrized model $f_\theta$ over a dataset $\mathcal{D} = \left\{ (x_i, y_i) \right\}_{i=1}^n$ using a differentiable loss function $\ell(f_\theta(x_i), y_i)$ measuring the performance of the model predictions vs. the ground truth label on each samples. Unless specified otherwise, the following experiments set the loss function to be the mean square error loss MSE$(x,y) = (x - y)^2$ to be compliant with the framework settings we've set to prove our main theoretical claims and results.

\paragraph{Gradient based optimization.}
In what follows, we investigate the difference in optimization dynamics for vanilla gradient descent (GD), mini-batch stochastic gradient descent (SGD) and their regularized analogs in minimizing the (regularized) empirical risk. We refer to GD as the classic first-order optimization algorithm that updates the parameters $\theta$ of the model to minimize the objective function $\hat{\mathcal{R}}_\mathcal{D}(f_\theta)$ or $\hat{\mathcal{R}}_\mathcal{D}^\lambda(f_\theta)$ and mini-batch SGD, as the variation of GD that approximates the gradient using a subset $\mathcal{B}_t$ (or mini-batch) of size $m$ sampled from the training data at each step.
\begin{align*}
\textrm{(GD)} \quad \theta_{t+1} = \theta_t - \eta \nabla_\theta \hat{\mathcal{R}}_\mathcal{D}(f_{\theta_t})
\quad \textrm{and} \quad
\textrm{(SGD)} \quad \theta_{t+1} = \theta_t - \eta \nabla_\theta \hat{\mathcal{R}}_{\mathcal{B}t}(f_{\theta_t})
\end{align*}
where $\eta > 0$ is the learning rate, and $\nabla_\theta \hat{\mathcal{R}}_\mathcal{D}(f_{\theta_t})$ denotes the gradient of the risk with respect to the parameters $\theta$ at step $t$ over the entire dataset, while $\hat{\mathcal{R}}_{\mathcal{B}t}(f_{\theta_t}) = \frac{1}{m}\sum_{(x_i, y_i) \in \mathcal{B}t}\ell(f_{\theta_t}(x_i), y_i)$ denotes the gradient of the risk over the random mini-batch $\mathcal{B}_t$. While these two optimizers look to be very similar, their trajectory in the loss landscape and the solutions they converge to are very different in nature. Apart from the optimization dynamic itself, using GD at each step is also computationally very intensive for large datasets and its common to use mini-batches to reduce the compute burden.

\subsection{Synthetic regression target functions.}

\paragraph{Target function with sparse support.}
The main claim of the paper is to demonstrate that GD and mini-batch SGD lead to very different solutions on $\W_1$. In particular, we want to show that mini-batch SGD shrinks the weights associated to the irrelevant components of inputs to zero while GD does not. In order to properly investigate the behavior over optimization trajectories, we need proper handle on irrelevant components, to be able to pinpoint and track weights interacting with those components. To do so, we decided to first start with handcrafted synthetic regression problems $y(\mathcal{X})$ that depends only on a subset of components of $\mathcal{X}$ we control and where we know the true support of the task.

The base learning task we decided to use as a starting point for our empirical experiment is hence;
\begin{align*}
y(\x) = g\left( \sum_{i=1}^{r} \alpha_i h_i(x_i) \right) \quad \x \in \mathbb{R}^d \quad r < d.
\end{align*}
for some feature functions $h_i : \mathbb{R} \mapsto \mathbb{R}$ acting on each of the first $r < d$ components of $\x$ and an outer function $g : \mathbb{R} \mapsto \mathbb{R}$ acting on the linear combinations of features of $\x$. Very clearly, in this setting, we know \textit{a priori} what is the true support of the function, namely the first $r$ components of $\x$ (given $\alpha_i \neq 0$ and $h_i$ is not the zero function). We expect a good network to focus on the signal of the first $r$ components only and disregard all the $(d-r)$ irrelevant components of the input— which effectively only bring noise to the inference problem. In particular, if we set $\alpha_i = 1$ and $h_i(x) = g(x) = x$ we get a simple linear target function of the first $r$ components of input $y(\x) = \sum_i^r x_i$ with true support being $\x[:r]$. We can extend to arbitrarily more sophisticated tasks with choices of $g(.), h_i(.), \alpha_i$.

The objective is now to train a range of deep neural network architectures using GD, SGD and their regularized analogs and investigate what are the solutions we get, and what is the dynamic of the \textit{a priori} irrelevant columns of $\W_1$ (those associated with $\x[r:]$) evolve over the optimization process.

\paragraph{Ablations.}
We decided to pick the following two classes of synthetic target functions. The linear function $y(\x) = \sum_i^r x_i$ and the sine non-linearity $y(\x) = \sin \left( \sum_i^r x_i \right)$. In what follows we performed extensive ablations on linear networks and ReLU networks of different sizes initialized with normal Kaiming He \cite{he2015delving}, and a range of hyperparameters. Same behavior occurs with other classes of targets.

\paragraph{Data generation.}
Once we design the target function $y(\mathcal{X})$ we generate dataset $\mathcal{D} = \left\{ (x_i, y_i) \right\}_{i=1}^n$ by first sampling a data matrix $\mathbf{X}$ consisting of $m = 5000$ samples each with $d=15$ dimensions from a standard normal distribution $\mathcal{N}(\mathbf{0},\mathbb{I}_d)$. The first $r=5$ dimensions are considered relevant. We center the irrelevant features (dimension $6$ to $15$) by subtracting their respective means, ensuring that they contribute no informative variance in the target $y(\x)$. We then create two sets of response variables $(\mathbf{Y}_1, \mathbf{Y}_2)$ generated by fast-forwarding data matrix $\mathbf{X}$ through our pre-designed targets and by adding and subtracting small Gaussian noise ($\epsilon$) respectively. This ensures that we get symmetric oscillations over all samples. The final dataset consists of the $n = 2m$ samples $([\mathbf{X}, \mathbf{X}], [\mathbf{Y}_1, \mathbf{Y}_2])$.

\subsection{Experiments for synthetic linear target function.}

\begin{figure}[ht]
\centering
\includegraphics[width=1.0
\textwidth]{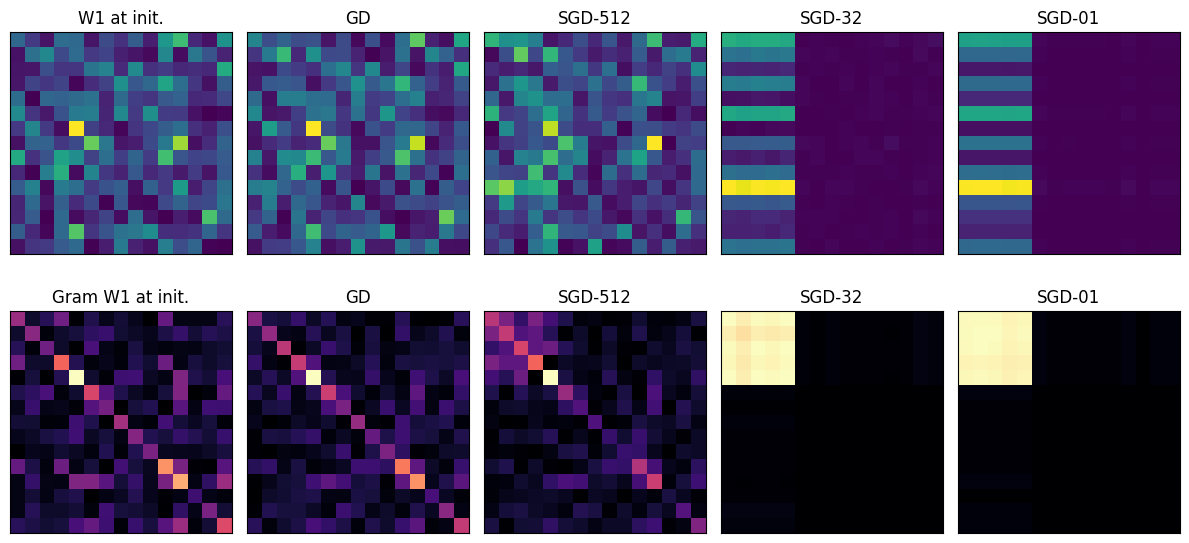}
\caption{
\textbf{Weights $\W_1$ and grams at initialization and convergence in a 3x15 MLP network.}
}
\label{fig:support:linear:gw}
\end{figure}

\paragraph{Convergence on $\W_1$.} In Fig.\ \ref{fig:support:linear:gw}, we illustrate the evolution of layer $\W_1$ and Gram matrices within a linear MLP network consisting of 3x15 hidden neurons, highlighting both initial configurations and patterns at convergence using GD and SGD with a range of batch sizes. The top row displays direct visualizations of the weight matrices, while the bottom row depicts the associated Gram matrices.

\begin{figure}[ht]
\centering
\includegraphics[width=0.32\textwidth]{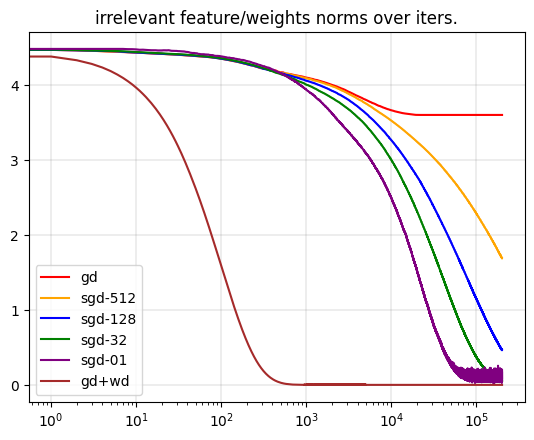}
\includegraphics[width=0.32\textwidth]{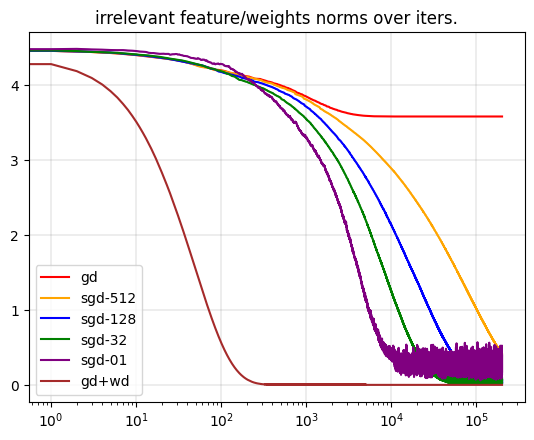}
\includegraphics[width=0.32\textwidth]{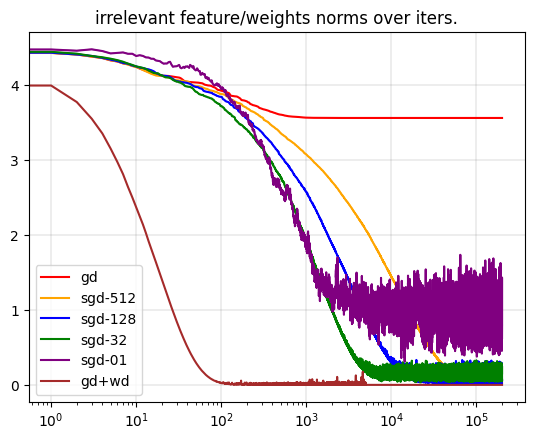}
\caption{
\textbf{Norm of irrelevant weights of $\W_1$ over time in a 3x15 Network.} $\eta = 0.1, 0.2, 0.5$.
}
\label{fig:support:linear:irelnorms}
\end{figure}

\paragraph{Irrelevant weights of $\W_1$.} In Fig.\ \ref{fig:support:linear:irelnorms}, we show the norm of the weights of $\W_1$ associated with irrelevant components of inputs for a range of batch size and learning rates. For mini-batches, we read the median trajectory over $10$ random trajectories (sampling order). We can clearly observe that GD firmly stops at some point, while SGD and GD with regularization keep shrinking those weights to zero. The speed at which SGD shrinks the weights is proportional to $\eta / b$ (step size / batch size).

\begin{figure}[ht]
\centering
\includegraphics[width=1.0\textwidth]{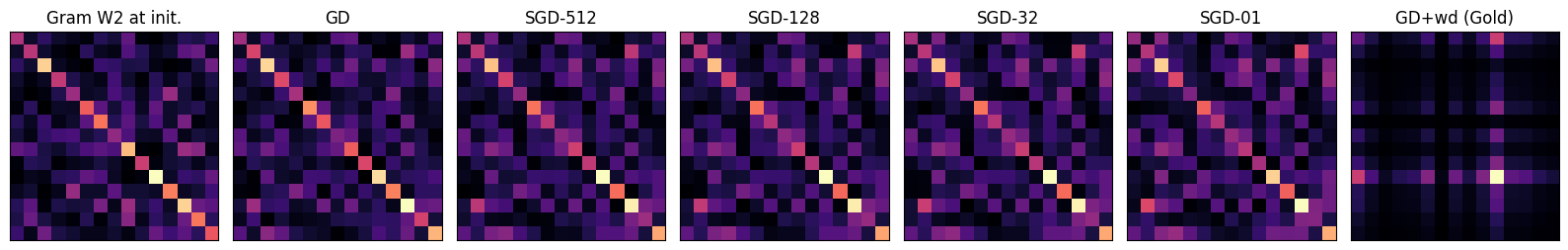}
\includegraphics[width=1.0\textwidth]{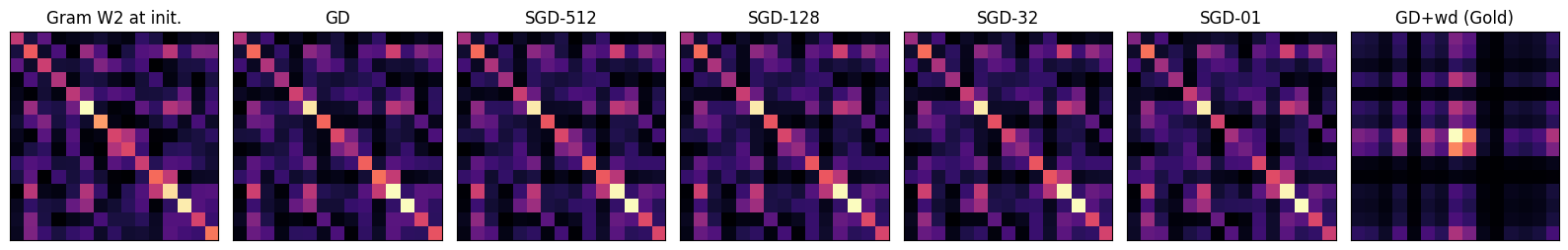}
\caption{
\textbf{Gram matrices of upstream weight layers $\W_2, \W_3$ at initialization and convergence.}
}
\label{fig:support:linear:upstream}
\end{figure}

\paragraph{Gram matrix of downstream layers.} We show that the SGD target support identification property that we observe in Fig.\ \ref{fig:support:linear:gw} does not carries over to downstream layers. In Fig.\ \ref{fig:support:linear:upstream} we show the Gram matrices of $\W_2, \W_3$ at initialization and convergence for a number of optimizers.

\subsection{Experiments for synthetic sine target function.}

\begin{figure}[ht]
\centering
\includegraphics[width=1.0
\textwidth]{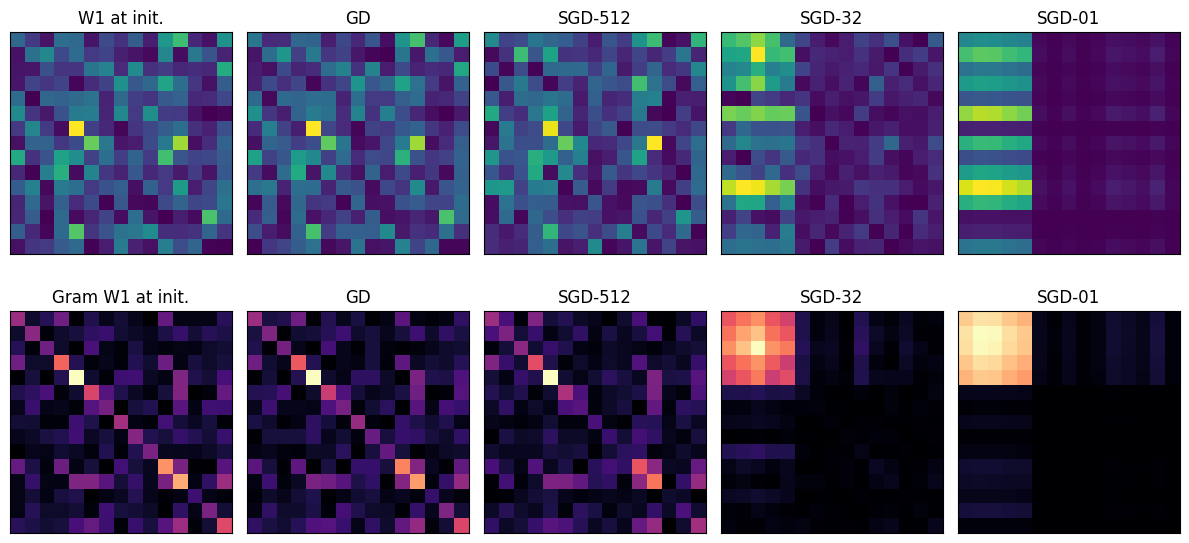}
\caption{
\textbf{Weights $\W_1$ and grams at initialization and convergence in a 3x15 MLP network.}
}
\label{fig:support:sine:gw}
\end{figure}

\paragraph{Convergence on $\W_1$.} In Fig.\ \ref{fig:support:sine:gw}, we illustrate the evolution of layer $\W_1$ and Gram matrices within a linear MLP network consisting of 3x15 hidden neurons, highlighting both initial configurations and patterns at convergence using GD and SGD with a range of batch sizes. The top row displays direct visualizations of the weight matrices, while the bottom row depicts the associated Gram matrices.

\begin{figure}[ht]
\centering
\includegraphics[width=0.32\textwidth]{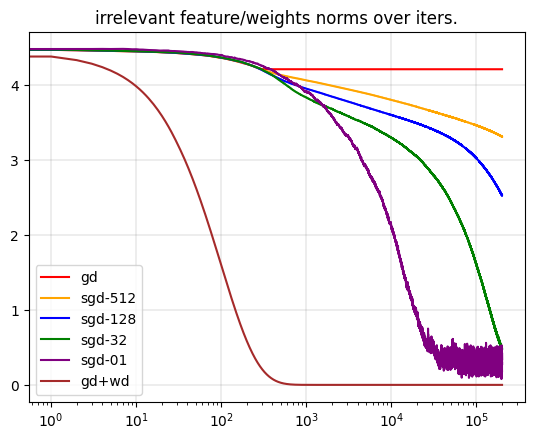}
\includegraphics[width=0.32\textwidth]{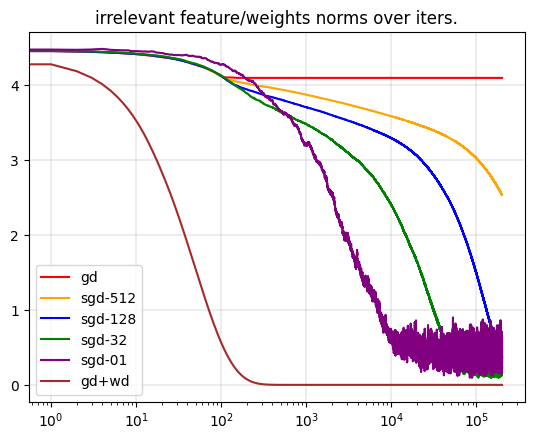}
\caption{
\textbf{Norm of irrelevant weights of $\W_1$ over time in a 3x15 Network.} $\eta = 0.1, 0.2$.
}
\label{fig:support:sine:irelnorms}
\end{figure}

\paragraph{Irrelevant weights of $\W_1$.} In Fig.\ \ref{fig:support:sine:irelnorms}, we show the norm of the weights of $\W_1$ associated with irrelevant components of inputs for a range of batch size and learning rates. For mini-batches, we read the median trajectory over $10$ random trajectories (sampling order). We can clearly observe that GD firmly stops at some point, while SGD and GD with regularization keep shrinking those weights to zero. The speed at which SGD shrinks the weights is proportional to $\eta / b$ (step size / batch size).

\paragraph{Initialization independent.} We run the same experiments over 4 different seeds.
\begin{figure}[ht]
\centering
\includegraphics[width=\textwidth]{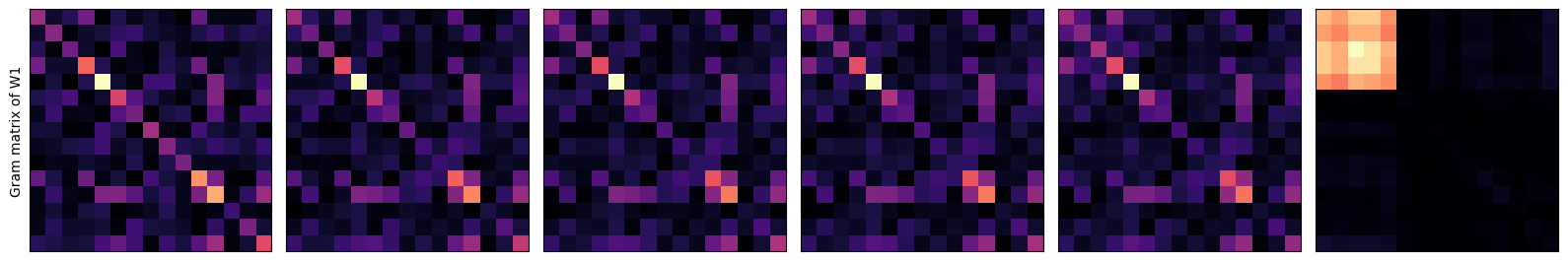}
\includegraphics[width=\textwidth]{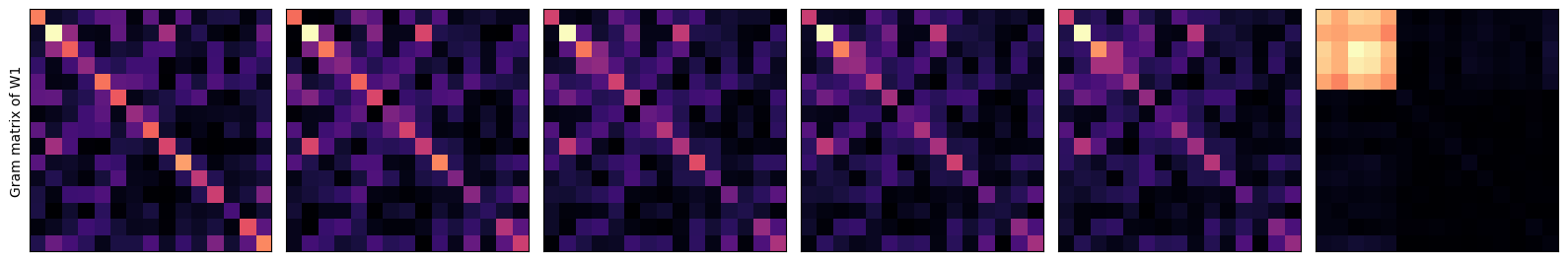}
\includegraphics[width=\textwidth]{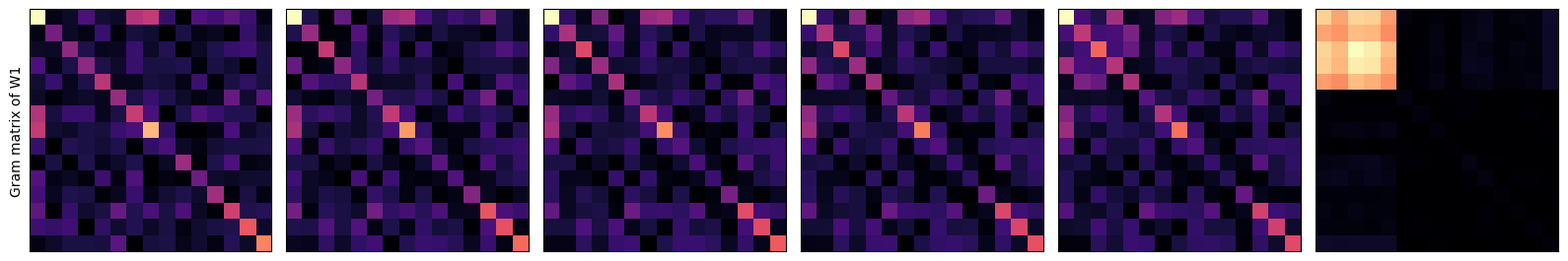}
\includegraphics[width=\textwidth]{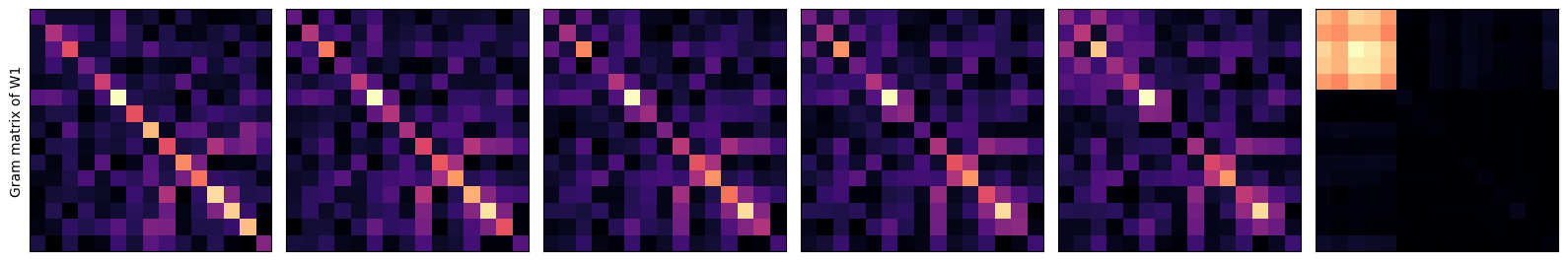}
\caption{\textbf{Weights $\W_1$ and Gram matrices at initialization and at convergence for several seeds.}}
\label{fig:support:sine:relu:gw}
\end{figure}

\paragraph{With ReLU Networks.} SGD with lower batch sizes picks up the support more confidently.
\begin{figure}[ht]
\centering
\includegraphics[width=\textwidth]{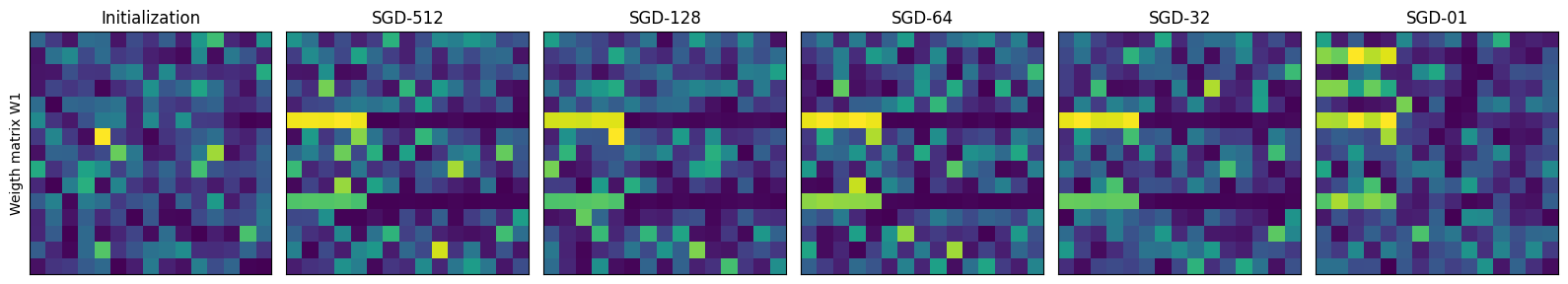}
\includegraphics[width=\textwidth]{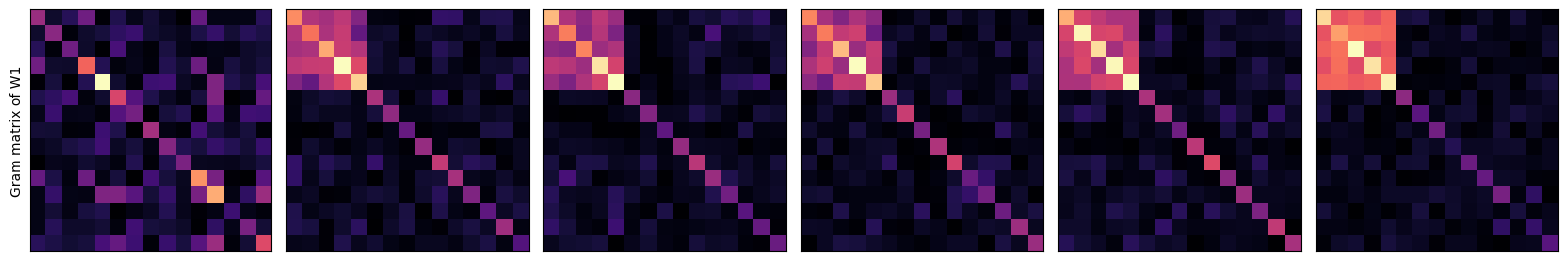}
\caption{\textbf{Weights $\W_1$ and Gram matrices at initialization and at convergence for ReLU.}}
\label{fig:support:sine:relu:gw_2}
\end{figure}

\section{Support Identification : CIFAR10 With ResNet16}
\label{appendix:support:cifar10}
While MNIST still fits pretty well with our standard settings, namely we have a handle on irrelevant components of inputs that are roughly speaking the corner of the input images, the situation is much less clear with images coming from CIFAR10 \cite{krizhevsky2009learning}, CIFAR100 \cite{krizhevsky2009learning}, and ImageNet \cite{deng2009imagenet} datasets.

\begin{figure}[ht]
\centering
\includegraphics[width=0.8\textwidth]{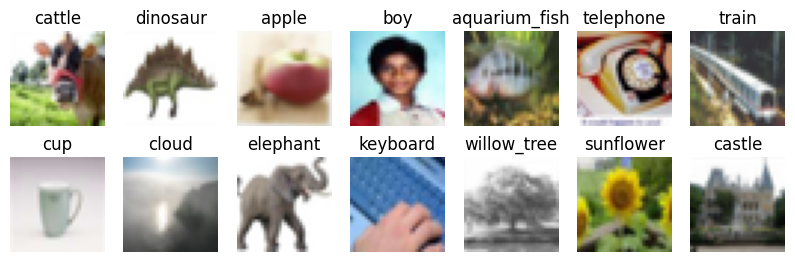}
\caption{A sample grid displaying a couple of images from the CIFAR100 dataset.}
\label{fig:vision:cifar100:grid}
\end{figure}

We consider a 11.1M parameters ResNet18 model pretrained on CIFAR10 achieving a 93.07\% accuracy (\url{https://github.com/huyvnphan/PyTorch_CIFAR10}). This model consists of \orange{$x$} residual blocks of convolutional layers and a fully connected linear map regressing over the features extracted by the upstream ResNet. We write $f(\x) = \mathcal{G} \circ \textrm{Res}(\x)$ where $\x$ is the (batch) of input images and where $\textrm{Res}(.)$ refers to the mapping of residual blocks. This pre-trained model is adapted by freezing all convolutional layers and replacing the top classification layer with a 3 layers MLP head, designed to only update during training, focusing on adapting high-level features extracted by the ResNet. The MLP head is initialized with Kaiming Uniform. The training is conducted solely on the MLP, utilizing the robust feature representations from the frozen ResNet18 base. We show that we can observe the same support identification property with SGD.

\paragraph{Finding the ground truth.} The challenge we have when dealing with an MLP trained on vector features extracted from a pretrained residual network $z \in \mathbb{R}^{512}$, is that we don't have a clear handle on which of those are irrelevant for the target function. To do so, we first train the model with an aggressive weight decay at convergence, to get to a model that performs equally well in accuracy over the test data, but has successfully identified the "ground truth" support.

\begin{figure}[ht]
\centering
\includegraphics[width=1.0\textwidth]{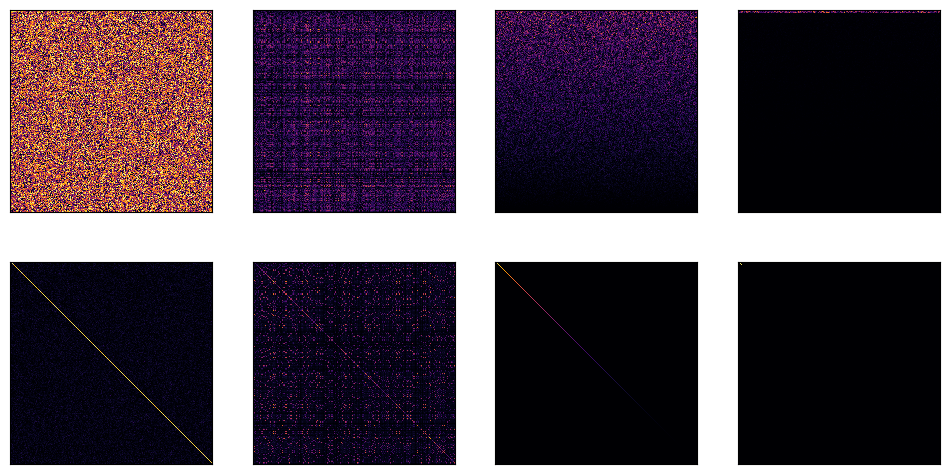}
\caption{Gram matrices of the first hidden layer of the MLP head, multiplied on the right by $V^\top$ where $V$ is derived from the SVD decomposition of $W$. On the left, the matrix at initialization; on the right, the matrix at convergence after training with SGD and weight decay. The visualization shows that the layer has identified a compact support in this new basis, indicating that many of the features from the ResNet contain (quasi) irrelevant information.}
\label{fig:head:cifar10:ground:truth}
\end{figure}

\begin{figure}[H]
\centering
\includegraphics[width=1.0\textwidth]{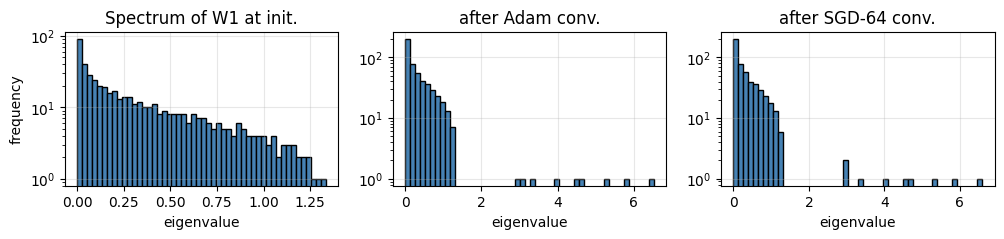}
\caption{\textbf{Eigenvalues of the first weigth layer of the MLP head on CIFAR10.}}
\label{fig:head:cifar10:ground:truth:histo:adam:sgd}
\end{figure}


\section{Theory: Linear Setting}
\label{setting_data}

\subsection{Data Assumptions for Linear Networks}
\paragraph{Assumption 1: Relevant and irrelevant features.}
Suppose we are given a dataset $\mathcal{D} = \left\{ x_i, y(x_i) : x_i \in \mathbb{R}^d \right\}_{i=1}^n$. 
Let $\Lambda_{y,x} = \E[yx^\top]$ be of rank $r$. Let $\bar{\mathcal{B}}$ be a basis of right eigenvectors for $\Lambda_{y,x}$ ordered by size of the corresponding eigenvalue. This means that on average $y$ depends only on the first $r$ vectors of the basis $\bar{ \mathcal{B}}$, we will call these the relevant components and the other $d-r$ the irrelevant components. 
Assume that the relevant and the irrelevant components are scorrelated, or more precisely $\E_{x \in \mathcal{D}}[x[i]x[j]]=0$ for all $i \leq r < j$.

\paragraph{Loss.}
We will train neural networks models denoted by $f_\theta(x)$, where $\theta$ is the vector of parameters and $x$ the input, on the dataset defined above with MSE loss. As follows
\[
\frac{1}{2n} \sum_{(x,y) \in \mathcal{D}} \mednorm{f_\theta(x) - y}_2^2.
\]
When we mention weight decay we will train instead on the Ridge penalized loss
\[
\frac{1}{2n} \sum_{(x,y) \in \mathcal{D}} \mednorm{f_\theta(x) - y}_2^2 \ + \ \frac{\lambda}{2}\mednorm{\theta}^2_2.
\]

\paragraph{Models.}
The models considered are linear and ReLU, diagonal or fully connected neural networks.

\paragraph{The Optimization Algorithms.}
Let $\eta>0$ be the learning rate or step size. We call gradient descent or full-batch gradient descent with step size $\eta$ started from $\theta$ the following iterative algorithm
\begin{equation}
\begin{split}
    \theta_0 \quad &= \quad \theta
    \\ 
    \theta_{k+1} \quad &= \quad 
    \theta_k \ - \ \eta \cdot \frac{1}{n} 
    \sum_{(x,y) \in \mathcal{D}} \nabla_\theta L\big(\theta, (x,y)\big)
    \qquad \text{for all } k \in \N.
\end{split}
\end{equation}
We define mini-batch SGD the same algorithm but where at the step $k$ the average is taken over $(x,y) \in B_k$ the $k$th batch of data.
\begin{equation}
\begin{split}
    \theta_0 \quad &= \quad \theta
    \\ 
    \theta_{k+1} \quad &= \quad 
    \theta_k \ - \ \eta \cdot \frac{1}{|B_k|} 
    \sum_{(x,y) \in B_k} \nabla_\theta L\big(\theta, (x,y)\big)
    \qquad \text{for all } k \in \N.
\end{split}
\end{equation}
We denote \textit{SGD without replacement} the version in which the batches are sampled without replacement from $\mathcal{D}$ until unseen data points in $\mathcal{D}$ are exhausted (this number of steps is called an epoch), then the sampling restarts.
We denote \textit{SGD with replacement} the one in which the batches are sampled with replacement, so they are independent uniformely sampled subsets of the dataset $\mathcal{D}$.
We call \textit{Online SGD} the one in which the batches are sampled independently from a distribution, i.e., there is no dataset, the algorithms sees new data at every step.

When we consider an algorithm \textit{with weight decay} we mean that there exists a value $\lambda_k > 0$ (often $\lambda_k = \lambda$ is constant) such that the $k$th optimization step is the one of the algorithm on the loss plus $\frac{\lambda_k}{2} \mednorm{\theta}_2^2$. For instance, GD with weight decay is
\begin{equation}
\begin{split}
    \theta_0 \quad &= \quad \theta
    \\ 
    \theta_{k+1} \quad &= \quad 
    \theta_k \ - \ \eta \cdot \frac{1}{n} 
    \sum_{(x,y) \in \mathcal{D}} \nabla_\theta L\big(\theta, (x,y)\big)
    \ - \ \eta \lambda_k \theta_k
    \qquad \text{for all } k \in \N.
\end{split}
\end{equation}

\paragraph{Change of basis and rotation invariance.}
Note that $\Lambda_{x,x} := \E_{x \in \mathcal{D}}[xx\top]$ is thus a symmetric PSD matrix that in the basis $\bar{ \mathcal{B}}$ satisfies $\Lambda_{x,x} [:r, r+1:] =0$.
Let us now take an orthonormal basis of eigenvalues for the covariance of the irrelevant components $\Lambda_{x,x}[r+1:, r+1:]$ and replace the irrelevant basis vectors of $\bar{ \mathcal{B}}$ with them in this new basis $\mathcal{B}$. In this new basis, the matrix $\Lambda_{x,x}[r+1:, r+1:]$ is diagonal and the matrix $\Lambda_{x,x}[:r, r+1:]$ is 0. Moreover, this is still an orthonormal basis of eigenvalues for $\Lambda_{y,x}$.
Consider without loss of generality $W_1$ to be in coordinate of $\mathcal{B}$ or analogously $(W_1)^{new} = W_1 \cdot M_{\mathcal{B}}^\top$. This does not impact the training and now we can properly study the dynamics of $W_1$ component by component where the components are meaningful. This is the case because gradient based methods are rotational invariant. If at the end of the training we multiply $W_1$ from the right with $M_{\mathcal{B}}$ we have the $W_1$ we would have trained on the data that did not change basis.

\subsection{Manipulating the (linear) Loss.}

\paragraph{Linear networks and loss manipulation.}
Consider now the model
\begin{equation}
\label{linear_nets}
f_\theta(v)   = Wv := 
W_L \cdot W_{D+1} \cdot \ldots
\cdot
W_1
\cdot v.
\end{equation}
Let $X$ and $Y$ be the data matrices, i.e., on every row we have the input $x_i$ and output $y_i$ of $i-th$ data point.
\begin{equation}
    L(\theta)
    \quad = \quad
    \frac{1}{n}
    \| XW^\top - Y \|_F^2
    \quad = \quad
    \frac{1}{n}
    \trace{[X W^\top - Y] [WX^\top - Y^\top]}.
\end{equation}
This trace can be rewritten as
\begin{equation}
    \frac{1}{n} \trace{X^\top X \cdot W^\top W}
    - \frac{2}{n} \trace{ X^\top Y \cdot W}
    + \frac{1}{n} \trace{Y^\top Y}.
\end{equation}
Now, noticing that the gradient based methods are rotation invariant, rereading this above in the basis $\mathcal{B}$, we have that this is equivalent to
\begin{equation}
\label{eq:manipulated_loss}
    \| W \cdot \Lambda_{x,x} - \Lambda_{y,x} \|_F^2 + R^2
\end{equation}
where $R^2 = \trace{\E_L[yy^\top]} - \trace{\Lambda_{y,x}\Lambda_{y,x}^\top}$.

From this representation, we can characterize the stationary points, indeed $\nabla_\theta L(\theta) = 0$ if and only if for all the components $\theta_k$ 
\begin{equation}
    [d_{\theta_k} W]^\top \quad 
    \cdot \quad 
    \left(
    W \Lambda_{x,x} - \Lambda_{y,x} \right)
    \quad = \quad
    0
    .
\end{equation}
Note that thus, the derivative in the parameter $W_1[j, i]$ of a layer of the output function $W$ is
\begin{equation}
    d_{W_1[j,i]} W
\quad = \quad
W_L \cdot W_{L-1} \cdot \ldots \cdot \cdot W_{2} \cdot E_{j,i}.
\end{equation}

\paragraph{The GD step for linear nets.}
Let us define $\tilde W = W_L \cdot W_{L-1} \cdot \ldots \cdot W_2$. The GD step on the weights $W_1[:,r+1:]$  relative to the irrelevant components in this case is
\begin{equation}
\label{GD:linear_step}
\begin{split}
    W_1[:, r+1:]
    \quad &= \quad 
    \ - \ \eta
    [\nabla_{W_1[j,r+1:]} W]^\top \quad 
    \cdot \quad 
    \left(
    W \Lambda_{x,x} - \Lambda_{y,x} \right)
    \\&=\quad
    \ - \ 
    \eta \cdot
    (\tilde W^\top
    \tilde W) W_1[:,r+1:] \Lambda_{x,x}[r+1:, r+1:]
    \\&\qquad \
    \ - \ 
    \eta \cdot
    (\tilde W^\top
    \tilde W) W_1[:,:r] \Lambda_{x,x}[r, r+1:]
    .
\end{split}
\end{equation}

\section{Dynamics of Gradient Descent}
\label{section:GD}

\subsection{Learning the Support}
In this section let $f_\theta(x) := W^{(L)} \cdot W^{(L-1)} \cdot \ldots \cdot W_1\cdot x$ be a linear fully connected neural network as in Eq.\ \ref{linear_nets}.
Let $\tilde W := W^{(L)} \cdot W^{(L-1)} \cdot \ldots \cdot W^{(2)} \in \R^{n_L \times n_2}$.
Consider throughout this section the stepsize $\eta > 0$ to be smaller than the instability threshold. I.e., smaller than the 2 divided by the eigest eigenvalue of the full-batch Hessian.

\begin{proposition}
    \label{prop:1}
    Assume we are in the setting above the decomposition in relevant and irrelevant components of \S \ref{setting_data}.
    Let $f(\theta, x)$ be a linear network.
    Then the absolute value of weights interacting with the irrelevant components $\textbf{w}$ is monotonically non-increasing during training with GD.
    Moreover, the GD iteration on $\textbf{w}$ acts as
    \[
    \textbf{w} \quad \curvearrowleft \quad
    \textbf{w} - \eta (\tilde W^\top \tilde W) \cdot \textbf{w} \cdot \Lambda_{x,x}.
    \]
\end{proposition}

\begin{remark}
    The result is even stronger, at every step we know exactly what weights are decrease and by how much. The speed on every weight is governed by the singular values of the following layers and the variance of the relative input component.
\end{remark}
Precisely we can conclude that the irrelevant components that are not spanned by the data, i.e., for which $\Lambda_{x,x} = 0$ remain exactly as at initialization. This is for instance the case of the pixels in the corners for MNIST when no data augmentation is used. 
\begin{proposition}
\label{prop:1a}
Let $f_\theta$ be linear. The components of the weights that interact with irrelevant components that are not spanned by the data remain untouched during the training.
\end{proposition}

\subsection{Not in the First Layer}

However, even though GD can learn the support, we can not make sense of the features and this does not happen in the first layer, but with uniform probability over any layer.
The idea is that every weigth in every layer is multiplied by all of the others, thus the one that is the smallest at initialization gets to zero first and the dynamics stops. This can happen with uniform probability at any layer.

\textbf{Proposition \ref{prop:GD}}(Negative result for GD, 1)\textbf{. }
\textit{Let $f_\theta$ be an either linear or ReLU diagonal network. Assume each weight is randomly initialized.
When trained with gradient descent or gradient flow the neural network learns the support, but the probability that it identifies it in the first layer is smaller than} $\mathrm{depth}^{-(d-r)}$.

Let us now quickly consider the case of a nonlinear activation $\sigma$, only for the sake of the next proposition.
\begin{proposition}[Negative result for GD, 2]
    \label{prop:4a}
    Assume we are in the setting above the decomposition in relevant and irrelevant components of \S \ref{setting_data}.
    Let $f_\theta$ be network with \emph{any non-linear activation}.
    Then the absolute value of weights interacting with the irrelevant components $\textbf{w}$ may never decrease.
\end{proposition}

\begin{remark}
    This means that neural networks with gradient descent learn the support and the same features that SGD would learn. However, those features are represented in a much less interpretable way. Indeed, we can not infer the support of these features by inspecting the first layer. We will see below that with SGD we can.
\end{remark}

\section{Mini-batch SGD}
\label{section:SGD}

In all the proofs above we have to swap the covariance matrix $\Lambda_{x,x}$ with the matrix $\Lambda_{x,x}^B$, that is the covariance of the inputs on the batch. On a single batch $B$ the eigenvalues and the eigenvectors of $\Lambda_{x,x}^B$ are not anymore the same of the covariance over the whole dataset. 
This implies that we have some movement along other directions too.
For SGD without replacement where the steps are not independent instead, the covariance matrices of the batches of the following steps will sum up to the full-batch covariance matrix. 
Precisely, if the batches are $B_1, B_2, \ldots, B_k$ we have
\[ \Lambda_{x,x}^{B_1} + \Lambda_{x,x}^{B_2} + \ldots + \Lambda_{x,x}^{B_k} \quad = \quad k \cdot \Lambda_{x,x}. \]
Thus if we analyze 2 or more steps together, we will have that at the first order we have the same movement but at the second or higher order there is a positive bias with respect to GD that depends on the variance of the matrices $xx^\top$, or better from the fourth order of the distribution. Indeed in the $i$th order effect of GD we have terms $\nabla^i_{x,x} = \E[xx^\top]^i$ while in the analogous term of SGD we have 
\[
\E[\underbrace{xx^\top \cdot xx^\top \cdot \ldots xx^\top}_{i\text{ times}}].
\]
This implies that the movement due to the higher order terms is bigger. In particular, since the sum of the covariances of the individual batches sum up to the full-batch one, the effect that we see is that if we moved of $\delta$ more in a direction in the following steps we will move of $-\delta$.
Thus essentially, if the values were shrunk as
$(1 - \eta c)$ for GD, now they get shrunk as
$(1 - \eta c - \eta \delta)$ and then as $(1 - \eta c + \eta \delta)$ and the resulting averaged rate is $(1 - \eta c - \eta^2 \delta^2)$.

\subsection{Learning the Support}

\begin{proposition}
    \label{prop:2}
    Assume we are in the setting above the decomposition in relevant and irrelevant components of \S \ref{setting_data}.
    Let $f(\theta, x)$ be a linear network.
    Then the absolute value of weights interacting with the irrelevant components $\textbf{w}$ is monotonically non-increasing during training with SGD. Moreover, their expected speed of convergence is always faster than the one they would have with GD. Precisely, there exists $\delta$ such that on average
    \[
    \textbf{w} \quad \curvearrowleft \quad
    \textbf{w} - \eta (\tilde W^\top \tilde W) \cdot \textbf{w} \cdot \Lambda_{x,x}
    - \frac{\eta^2}{(batch\ size)^2} (\tilde W^\top \tilde W)^2 \cdot \textbf{w} \cdot \delta^2
    .
    \]
\end{proposition}

For SGD and GD to converge it is needed that the step size, along the trajectory is smaller than as 2 divided by the highest eigenvalue of the Hessian of the loss. Since on the weights associated with the irrelevant components converge to zero and are non-increasing, we just need to bound this at initialization
\[
\eta_{\max} \quad = \quad 2/\lambda_{\max}(Hessian\ at\ initialization).
\]

This averaged gain in speed may not seem such a big difference. However, in case $c \to 0$ converges to 0 and $\delta$ does not, the difference becomes huge. In this case, eventually, we have no convergence for GD and we have \emph{exponential} convergence for SGD.
This scenario actually happens in practice in some cases. As an example let us take a diagonal linear network in the setting above.
\begin{proposition}
\label{prop:2b}
    Let $f_\theta$ be linear diagonal, assume the setting above. Then SGD with replacement learns the support in the first layer, no matter the initialization, if during the training dynamics the oscillations on the following quantity is lower bounded by a constant $c > 0$.
    \[
    \E_{B \subseteq \mathcal{D}}\mednorm{\E_{(x,y) \in B}\left[
    f_\theta(x) x^\top - yx^\top \right]}_2^2 
    \quad \geq \quad c > 0
    .
    \]
\end{proposition}

In particular, note that in the case of diagonal networks, on the component $j$ for which $\E_{\mathcal{D}}\big[\x[j]\x[j]\big] > 0$, the fact that the data is misspecified $R_j^2 := \E\big[y[j]^2\big] - \E\big[y[j] \x[j]\big]\E\big[\x[j]^2\big]^{-1}\E\big[y[j] \x[j]\big] > 0$ implies that for any choice of parameters the residuals on that component are generally non-zero
\begin{lemma}
\label{lemma:residuals}
    Assume we have any model that is linear in the data. Let $B\subseteq \mathcal{D}$ of size $b$ be a uniformly sampled batch of data points $(\x,y) \in \mathcal{D} \subseteq \R^2$. Assume $\mathcal{R} := \E\big[y^2 \x^2\big] - \E\big[y \x\big]^2 > 0$.
    Then 
    \[
    \E_{\mathcal{D}}\left[ \epsilon_\x^2 \right] - \E\big[\epsilon_\x\big]^2 
    \quad \geq \quad
    \frac{\mathcal{R}}{b}.
    \]
\end{lemma}

\textbf{Proposition \ref{prop:SGD}.}
\textit{Assume that $\E_{\mathcal{D}}\big[\x[j]^2\big]>0$, $\E_{\mathcal{D}}\big[y[j] \x[j]\big]=0$ and that we have $R^2 := \E\big[y[j]^2\big] - \E\big[y[j] \x[j]\big]\E\big[\x[j]^2\big]^{-1}\E\big[y[j] \x[j]\big] > 0$, i.e., the data is mispecified. Let $f_\theta$ be a diagonal linear network. Assume the step size $0 < \eta < \eta_{\max}$.
When trained with either SGD with or without replacement, the neural network learns the support in the first layer with high probability.
}

Note that Lemma \ref{lemma:2} proves Proposition \ref{prop:2b} and Lemma \ref{lemma:residuals} and Lemma \ref{lemma:2} prove Theorem \ref{theo:time}.

\section{Weight Decay}

Proposition \ref{prop:need4wd} is implied by the two following propositions.
For SGD and GD to converge it is needed that the step size, along the trajectory is smaller than as 2 divided by the highest eigenvalue of the Hessian of the loss, minus $\lambda$. Since on the irrelevant components 
\begin{proposition}
    \label{prop:3}
    Assume we are in the setting above the decomposition in relevant and irrelevant components of \S \ref{setting_data}.
    Let $f(\theta, x)$ be linear network as in Eq.\ \ref{linear_nets}. Assume that the step size $\eta$ is smaller than the instability threshold.
    Then the weights interacting with the irrelevant components $W_1[:, r+1:]$ converge to zero exponentially when using either SGD or GD with $\lambda-$weight decay.
\end{proposition}

\begin{proposition}
    \label{prop:3a}
    Assume we are in the setting above the decomposition in relevant and irrelevant components of \S \ref{setting_data}.
    Let $f(\theta, x)$ be ReLU network as in Eq.\ \ref{linear_nets}. Assume that the step size $\eta$ is smaller than the instability threshold.
    Then the weights interacting with the irrelevant components $W_1[:, r+1:]$ converge to zero exponentially when using any SGD with batch size 1 and with $\lambda-$weight decay.
\end{proposition}

\section{ReLU Setting}
\label{section:ReLU_app}

\paragraph{Representation of ReLU networks.}
Consider now the model
\begin{equation}
\label{nonlinear_network}
f_\theta(v) := 
W_L \cdot \sigma_{L-1} \circ W_{L-1} \cdot \ldots
\cdot
\sigma_1 \circ
W_1
\cdot v
\end{equation}
where $\sigma_i(x) = \max\{x,0\}$ is the ReLU activation function. 
Note that on every input $x$ there exists a linear network 
\begin{equation}
\tilde f_\theta^x(v) := 
\tilde W_L^x \cdot \tilde W_{L-1}^x \cdot \ldots
\cdot
\tilde W_1^x
\cdot v
\end{equation}
such that $f_\theta(x) = \tilde f_\theta^x(x)$.
Precisely the weights of the neural network can be repsresented as
\begin{equation}
\tilde W_i^x \quad = \quad D_i^x \cdot W_i D_{i-1}^x
\end{equation}
where $D_i$ is the diagonal matrix with the derivatives of the activation evaluated at the previous layers after the input $x$ on every diagonal component, i.e., with either 0 or 1 on the diagonal depending if the corresponding neuron of that layer is activated or not.

\paragraph{Derivative in the first layer.}
Note that
\begin{equation}
    \nabla_{W_1[i,j]} f_\theta(x)  
    \quad = \quad 
    W_L \cdot D_{L-1}(v) \cdot W_{L-1} \cdot \ldots
\cdot
D_1(v) \cdot
E_{i,j}
\cdot v
\end{equation}
where $D_i(v)$ is the diagonal matrix with the derivatives of the activation evaluated at the previous layers after the input $v$.
This means that if the gradient of the analogous linear network (as Eq.\ \ref{linear_nets}) is 
\begin{equation}
    \nabla_{W_1[i,j]} f_\theta(v)  
    \sum_{(i_L, \ldots i_3, i)} W_L[:, i_L] \prod_{j=L-1}^3 W_j[i_{j+1}, i_j] W_2[i_3, i] v_j
\end{equation}
the gradient of the nonlinear network at the same parameters is
\begin{equation}
    \nabla_{W_1[i,j]} f_\theta(v)  
    \sum_{(i_L, \ldots i_3, i)} W_L[:, i_L]\prod_{j=L-1}^3 W_j[i_{j+1}, i_j] W_2[i_3, i] 
    \ \cdot \ 
    \prod_{j=L-1}^2  D_{j}^x [i_{j+1},i_{j+1}]
    D_{j}^x[i,i] v_j
\end{equation}
where $\prod_{j=L-1}^2  D_{j}^x [i_{j+1},i_{j+1}]
    D_{j}^x[i,i] v_j$ are non-negative numbers.
More precisely, the dynamics only affect a precise sub-network for every datapoint. In particular, the step of the gradient on every datapoint is the same as the loss for linear networks. However, the model changes for every datapoint.

\paragraph{The setting: mispecification for ReLU networks.}
This implies that for a component $x[j]$ of the inputs satisfying $\E_{\mathcal{D}}[yx[j]] = 0$ on the dataset is not enough to be irrelevant. Precisely, in the case of ReLU there are no misspecified models. 

Thus the definition of an irrelevant component $j$ in this case is either
\begin{enumerate}
    \item $y$ is independent from $x[j]$. There exists a function $\phi$ such that $d_{x[j]} \phi(x) = 0$ for or all $x \in \R^d$ and we can represent the data as $\mathcal{D} = \{(x_i,\phi(x_i)\}_{i=1}^n$.
    \item Not spanned by the data: $x[j] = 0$ for all $(x,y) \in \mathcal{D}$.
\end{enumerate}
In this case, $\phi$ will be the learned function restricted to that component, i.e., if the component can be seen as irrelevant it will be treated as irrelevant.

\paragraph{Training the first layer.}
Let us work with a shallow ReLU network as in \S \ref{section:ReLU}
The effect of an SGD iteration on the batch $B$ on the weights $W_1[:, r+1:]$ corresponding to the irrelevant components can be rewritten as
\begin{equation*}
\begin{split}
    - \ 
    \frac{\eta}{n} \sum_{(x,y) \in B}
    \underbrace{(D^x W_2^\top W_2 D^x)}_{PSD} \cdot W_1[:,r+1:] \cdot \underbrace{x[r+1:]x[r+1:]^\top}_{PSD} 
    \qquad &\leftarrow \qquad \substack{\text{non-increasing on the}\\ \text{irrelevant weights}}
    \\- \ 
    \frac{\eta}{n} \sum_{(x,y) \in B}
    \underbrace{D^x W_2^\top}_{\text{a matrix}} \cdot \underbrace{\big( W_2 D^x W_1[:,:r] \cdot x[:r] - y\big)}_{\text{residuals on the relevant input features}}
    \cdot \underbrace{x[r+1:]^\top}_{\text{irrelevant}}
    \qquad \ \ \ &\leftarrow \qquad \substack{\text{unpredictable direction.}}
\end{split}
\end{equation*}
In the linear case we could conclude that the second line is $0$ in expectation over the data or that it induces benign oscillations for SGD without replacement. In the case of ReLU we can not as $W_2 D^x$, unlike $W_2$, is not the same on every datapoint. That said, this suggests that if the neural network has converged on the relevant input features, continuing to train is benign for our purpose.
\begin{proposition}
    Assume the function is learned on the relevant components, i.e., for all $x$ we have $W_L D^x_L W_{L-1} \cdot \ldots \cdot W_2 D^x_1 W_1[:,:r] \cdot x[:r] = y$. Then
    a step of either GD or mini-batch SGD is non-increasing on the weights of the first layer corresponding to the irrelevant components.
    More precisely, it is stationary on the rows orthogonal to $x[r+1:]$ and on the neurons that belong to the kernel of $W_L D^x_L W_{L-1} \cdot \ldots \cdot W_2 D^x_1$, it is decreasing otherwise.
\end{proposition}
This result, unfortunately, is not strong enough to conclude that we obtain convergence, as on the next step $W_2$ and $D^x$ may be different and so the relevant components may induce new changes. However, this is strong enough to say that if we proceed to train only the first layer, it learns the support on \emph{some} of the rows.

But assume we are training all the layers, than the solution is that either $D^x$ zeroes out that component, $W_2$ zeroes out that component, that component is perpendicular to $x[r+1:]$, or that component is zero. Indeed, $W_2$ and $W_1$ would change, than $W_2$ would change back due to the disalignment on $W_1[:,:r]$ and $W_1[:,r+1:]$ would have had 2 shrinking steps.

Moreover, we can anyways conclude, for instance, that if the convergence on the relevant features happens exponentially fast, then throughout the whole training the contribution of the second term is bounded, while the first one is not in principle. Thus if we run SGD for very long the first component actually shrinks those weights.

\section{Failure of the GD Argument for ReLU}
\label{section:ReLU_failure}
The GD step on the weights $W_1[:,r+1:]$ relative to the irrelevant components in case of ReLU networks is
\begin{equation}
\begin{split}
    W_1[:, r+1:]
    \quad &= \quad 
    \ - \ 
    \frac{\eta}{n} \sum_{(x,y) \in \mathcal{D}}
    \nabla_{W_1[:,r+1:]} f(\theta, x)^\top \
    \cdot \
    [f(\theta, x)-y]
    \\&=\quad
    \ - \ 
    \frac{\eta}{n} \sum_{(x,y) \in \mathcal{D}}
    (\tilde W(x)^\top
    \tilde W(x)) W_1[:,r+1:] \cdot x[r+1:]x[r+1:]^\top
    \\&\qquad \
    \ - \ 
    \frac{\eta}{n} \sum_{(x,y) \in \mathcal{D}}
    (\tilde W(x)^\top
    \tilde W(x)) W_1[:,:r] \cdot x[:r]x[r+1:]^\top
    .
\end{split}
\end{equation}
While the term in line 2 of the equation above still behaves well and implies a shrinking of the irrelevant weights, now the situation is a bit more problematic.
Indeed, $\tilde W(x)^\top \tilde W(x)$ changes for every datapoint. This implies that is there is a datapoint on which $\cdot x[:r]x[r+1:]^\top$ is negative and one on which it is positive, the two things do not cancel out and the term in line 3 of the equation above is not zero.

\paragraph{Counterexample to the argument for linear networks.}
    Assume, that $0 < x_1[r+1:]=x_2[r+1:] \ll 1$ is made of every small positive numbers for two datapoints $x_1$ and $x_2$ for which $x_1[:r]= - x_2[:r] \gg 1$ and is actually very big in size. Assume the weights relative to all the components in the first layer are of comparable size.
    This implies that all the neurons that are on for $x_1$ are off for $x_2$ and viceversa, as the sign of $W_1[j, :]x$ depends on the relevant components of $x$ in this case where they are much bigger than the irrelevant ones.
    
    Assume the target function is the absolute value of the sum of the relevant components. 
    This implies that the activated weights in the following layers of the networks are converging to essentially one.
    This means that assuming the network learns, at some point during the training the activated weights in the following layers are positive (as they are converging to 1) and the step will be
    \begin{equation}
\begin{split}
    W_1[:, r+1:]
    \quad &= \quad 
    \underbrace{- \ 
    \frac{\eta}{n} \sum_{(x,y) \in \mathcal{D}}
    (\tilde W(x)^\top
    \tilde W(x)) W_1[:,r+1:] \cdot x[r+1:]x[r+1:]^\top}_{\text{Wery small although it shrinks}}
    \\&\quad \underbrace{- \ 
    \frac{\eta}{2}
    (\tilde W(x_1)^\top
    \tilde W(x_1)) W_1[:,:r] \cdot x_1[:r]x_1[r+1:]^\top}_{\text{big positive}}
    \\&\quad \underbrace{- \ 
    \frac{\eta}{2}
    (\tilde W(x_2)^\top
    \tilde W(x_2)) W_1[:,:r] \cdot x_2[:r]x_2[r+1:]^\top}_{\text{big negative}}
    .
\end{split}
\end{equation}
This implies that on the rows that are activated by $x_1$ we are effectively learning the support. But we are effectively increasing by a similar size the rows activated by $x_2$.

\paragraph{Practical example.}
    Assume we have the following dataset $\mathcal{D} = \{ (x_i, x_i[1])\}_{i=1}^n \subseteq \R^d \times \R$. 
    Let the model be a 2 layer neural network with width $3$.
    Then for every $\alpha$, for every $v \in \R^{d-1}$ any network of the form
    \[
    W_2 = (-1, \alpha, 1-\alpha)
    , \quad 
    W_1 = \begin{pmatrix}
        -1 & 0 \\
        1 & (\alpha-1) v \\
        1 & \alpha v
    \end{pmatrix}.
    \]
    Then the represented function is
    \begin{equation*}
    \begin{split}
        -\max\{-x[1], 0\} + \max\left\{x[1] + (\alpha(\alpha-1)+(1-\alpha)\alpha) \langle v, x[2:] \rangle  \right\}
        \quad = \quad x[1]
    \end{split}
    \end{equation*}
    whenever both $|1-\alpha|$ and $|\alpha|$ multiplied by $|\langle v, x[2:] \rangle|$ on the data is smaller than $x[1]$ for all the datapoints in which $x[1] \geq 0$.
    This implies that if on the datasets the only entry with $x[1]=0$ is the vector $0$ (we just need something weaker than this). We can always find a $v\neq 0$ and $\alpha \in (0,1)$ such that the network exactly represents the function but it does not learn the support in the first layer, although it \textbf{approximately} does if $v$ is small in norm, and the approximation error is cut out by the second layer.

\paragraph{Other practical example.}
If one neuron of the first layer is off for all the datapoints on the training set at some point on the training, then the first weights are free to be any vector.
However, one could argue that the dynamics and the function represented are the same of a smaller network without that neuron.

\section{Proofs}
\label{section:proof}

\subsection{Proof of Proposition \ref{prop:1}, Proposition \ref{prop:1a} and Proposition \ref{prop:4a}}

Note that Appendix \ref{setting_data}, or better, Assumption \ref{ass:1} implies that 
$\Lambda_{x,x}[r, r + 1 :] = 0$. Plugging this into
Eq.\ \ref{GD:linear_step} implies that the GD step on
$W_1[:,r+1:]$ is
\begin{equation}
\label{eq:proof_1}
\begin{split}
   \!\!\!\! W_1[:,r+1:]
    \quad &= \quad 
    W_1[:,r+1:]
    \ - \
    \eta \cdot
    (\tilde W^\top
    \tilde W) W_1[:,r+1:] \Lambda_{x,x}[r+1:, r+1:]
    .
\end{split}
\end{equation}
This implies that opportunely changing orthonormal basis from left and right in a way that both $\tilde W^\top
    \tilde W$ and $\Lambda_{x,x}[r+1:, r+1:]$ are diagonal matrices with entries $D_W[i,i]$ and $D_x[j,j]$, the component $\bar W_1[i, j]$ of $W_1[:,r+1:]$ in the new basis becomes
\begin{equation}
\label{eq:GD_decreases_proof}
\begin{split}
    (1 - \eta D_W[i,i] D_x[j,j]) \bar W_1[i, j]
    .
\end{split}
\end{equation}
In particular, we have that the component that decreases the most reduces as $(1 - \eta \lambda_{\max}(\tilde W^\top \tilde W) \lambda_{\max}\Lambda_{x,x}[r+1:, r+1:] )$ and the components for which either the column corresponds to the irrelevant input variables without noise or the row to the kernel of $\tilde W$ remain untouched.
This, in particular, implies that as long as the span of the columns of $W_1[:,r+1:]$ is not contained in the Kernel of either $\tilde W$ or $\Lambda_{x,x}$ the frobenius norm decreases. For the other it does not move.
This concludes the proof of Proposition \ref{prop:1} and Proposition \ref{prop:1a}.

Proposition \ref{prop:4a} is proved by the counterexample in \S \ref{section:ReLU_failure}.
To be more detailed, we are working with $f_\theta(x) := W^{(L)} \cdot \sigma \circ W^{(L-1)} \cdot \ldots \cdot \sigma \circ W_1\cdot x$ be a linear fully connected neural network as in Eq.\ \ref{linear_nets}.
Let $\tilde W(x) := W^{(L)} \cdot D^x_L \cdot W^{(L-1)} \cdot \ldots \cdot D^x_2 \cdot W^{(2)} \cdot D^x_1 \in \R^{n_L \times n_2}$, where $D^x_l$ is the diagonal matrix with entries derivatives of $\sigma$ evaluated in the preactivations of the corresponding neuron. 
Note that if $\sigma$ is not the identity, Eq.\ \ref{eq:proof_1} is not true, indeed,  we have have
\begin{equation}
\begin{split}
   \!\!\!\! W_1[:,r+1:]
    \quad &= \quad 
    W_1[:,r+1:]
    \ - \
    \eta \frac{1}{n}\sum_{(x,y) \in \mathcal{D}} \tilde W^\top(x) [f_\theta(x)-y] x[r+1:]^\top
    .
\end{split}
\end{equation}
This update can be rewritten as
\begin{equation}
\begin{split}
    - \
    \eta \cdot \underbrace{\frac{1}{n}\sum_{(x,y) \in \mathcal{D}} \tilde W^\top(x) f_\theta(x) x[r+1:]^\top}_{\substack{\text{weighted covariance between}\\ \text{model-output and irrelevant inputs.}}}
    \ + \
    \eta \cdot \underbrace{\frac{1}{n}\sum_{(x,y) \in \mathcal{D}} \tilde W^\top(x) yx[r+1:]^\top}_{\substack{\text{weighted covariance between}\\ \text{data-output and irrelevant inputs.}}}
    .
\end{split}
\end{equation}
Here the weights are given by the model and they depend on $x$, so this is generally intractable. Under the assumption that $x[r+1:]$ are all unspanned this update is always zero. We can proceed splitting this dynamics expanding in Taylor in the input $f_\theta(x)$ and seeing what the second step is. However, in the general case, even with non-decreasing activation we can not conclude.
See the discussion in \S \ref{section:ReLU} as counterexample.

\subsection{Proof of Proposition \ref{prop:GD}}
Let us restrict WLOG to the subnetwork interacting with an irrelevant component $j$. This is of the form
\[
f_\theta(x)[j] \quad = \quad  \prod_{i=1}^L W_i[j,j] x[j].
\]
Then for all $h$ we have that 
\[
\dot W_h[j,j] \ = \ - \frac{1}{n} \sum_{x \in \mathcal{D}} 
\left(\prod_{i\neq h } W_i[j,j] x[j] \ \cdot \ \prod_{i=1}^L W_i[j,j] x[j] \right) \ = \ - \left( \prod_{i\neq h } W_i[j,j] \right)^2 \E_{\mathcal{D}}[x[j]^2] \cdot W_h[j,j].
\]
Thus all of the weights are moving towards 0 both with GD and gradient flow.
Almost surely one of them, in the layer $h$ is smaller than any other at initialization. This implies that the quantity
\[
\left( \prod_{i\neq h } W_i[j,j] \right)^2
\]
is bigger on that weight than on any other and that at the next step that one will still be the smallest and its rate of shrinking was bigger than any other. Applying induction on this argument establishes that eventually that one approaches zero and the other stabilize. Or better there exists a constant $c$ such that all the other are always bigger than that constant and this implies that $W_h[j,j]$ converges faster than exponentially with rate $(1 - \eta c^{2L-2}\E_{\mathcal{D}}[x[j]^2])$.
The element that was the smallest at initialization is any of them with uniform probability over the layers. There is one dynamic of this kind for every irrelevant component. 
Next note that in the case of ReLU,  this is what happens in the best case scenario only, in principle it may be that some neurons are completely off and so the weight matrices never learn that the corresponding component is unrelevant and the support is chosen by the activation.
This concludes the proof of Proposition \ref{prop:GD}.

\subsection{Proof of Proposition \ref{prop:2}}

Eq.\ \ref{GD:linear_step} and Appendix \ref{setting_data} implies that the GD step on
$W_1[:,r+1:]$ is
\begin{equation}
\begin{split}
    W_1[:,r+1:]
    \quad &= \quad 
    W_1[:,r+1:]
    \ - \
    \eta \cdot
    (\tilde W^\top
    \tilde W) W_1[:,r+1:] \Lambda_{x,x}^B[r+1:, r+1:]
    \\&\qquad \
    \ - \ 
    \eta \cdot
    (\tilde W^\top
    \tilde W) W_1[:,:r] \Lambda_{x,x}^B[r, r+1:]
    .
\end{split}
\end{equation}
Where the matrix $\Lambda_{x,x}^B$ is the covariance of the inputs on the batch. On a single batch the eigenvalues are not anymore the same of the covariance over the whole dataset. This implies that we have some movement along other directions too.
For SGD without replacement where the steps are not independent instead, the covariance matrices of the batches of the following steps will sum up to the full-batch covariance matrix. 
Precisely, if the batches are $B_1, B_2, \ldots, B_k$ we have
\[ \Lambda_{x,x}^{B_1} + \Lambda_{x,x}^{B_2} + \ldots + \Lambda_{x,x}^{B_k} \quad = \quad k \cdot \Lambda_{x,x}. \]
Thus if we analyze 2 or more steps together, we will have that at the first order we have the same movement but at the second or higher order there is a positive bias with respect to GD that depends on the variance of the matrices $xx^\top$, or better from the fourth order of the distribution. Indeed in the $i$th order effect of GD we have terms $\nabla^i_{x,x} = \E[xx^\top]^i$ while in the analogous term of SGD we have 
\[
\E[\underbrace{xx^\top \cdot xx^\top \cdot \ldots xx^\top}_{i\text{ times}}]
\]
That said, for SGD with replacement, where the batches are sampled independently, these covariance matrices are independent random matrices center on the full-batch covariance matrix. 
This implies that 
For SGD without replacement where the steps are not independent instead, the covariance matrices of the batches of the following steps will sum up to the full-batch covariance matrix. 

This implies that opportunely changing orthonormal basis from left and right in a way that both $\tilde W^\top
    \tilde W$ and $\Lambda_{x,x}[r+1:, r+1:]$ are diagonal matrices with entries $D_W[i,i]$ and $D_x[j,j]$, the component $\bar W_1[i, j]$ of $W_1[:,r+1:]$ in the new basis becomes
\begin{equation}
\begin{split}
    (1 - \eta D_W[i,i] D_x[j,j]) \bar W_1[i, j].
    .
\end{split}
\end{equation}
In particular, we have that the component that decreases the most reduces as $(1 - \eta \lambda_{\max}(\tilde W^\top \tilde W) \lambda_{\max}\Lambda_{x,x}[r+1:, r+1:] )$ and the components for which either the column corresponds to the irrelevant input variables without noise or the row to the kernel of $\tilde W$ remain untouched.

This implies that as long as the span of the columns of $W_1[:,r+1:]$ is not contained in the Kernel of $\tilde W$ the frobenius norm decreases. 

Denote by $x = \x[:r]$ the vector of the relevant components and $z= \x[r+1:]$ the one of the irrelevant components.
Denote for every batch $B \subseteq \mathcal{D}$ of data, the following submatrices as 
\[
\Lambda_{\x,\x}^B 
\quad :=\quad
\begin{pmatrix}
    \Lambda_{x,x}^B & \Lambda_{x,z}^B \\
    \Lambda_{x,z}^B & \Lambda_{z,z}^B 
\end{pmatrix}
\quad :=\quad
\frac{1}{|B|} \sum_{\x \in B} \x^\top\x 
\quad =\quad
\frac{1}{|B|} \sum_{(x,z) \in B} (x,z)^\top(x,z).
\]

Note that on one data point $\x, y(\x) = (x,z), Ax$ the residuals can be decoupled as
\begin{equation}
    W_2W_1 \x - y(\x)
    \quad = \quad
    \underbrace{(W_2W_1[:,:r] - A) x}_{function}
    \quad + \quad
    \underbrace{W_2W_1[:,r+1:] z}_{irrelevant}.
\end{equation}

Here, the step size $\eta>0$ is positive, the matrix $\mathrm{P}:= W_2^\top W_2$ is symmetric positive semidefinite, and $\Lambda_{z,z}^B$ is positive for every batch and it is centered around $\Lambda_{z,z}^D$. Thus what happens is that for full-batch GD the update on the weights of the first layer interacting with the irrelevant components is exactly 
\begin{equation}
\label{eq:GD}
\mathbf{W_1[:, r+1:]}
\quad - \quad  \underbrace{\eta \cdot
    \mathrm{P}}_{PSD} \ \cdot \ \mathbf{W_1[:, r+1:]} \ \cdot \
    \underbrace{\Lambda_{z,z}^\mathcal{D}}_{PSD}.
\end{equation}
In the case of mini-batch SGD, instead the step on the weights of the first layer interacting on the irrelevant components can be rewritten as
\begin{equation}
\label{eq:SGD}
\begin{split}
    \mathbf{W_1[:, r+1:]}
    \quad
    \underbrace{- \quad  \eta \cdot
    \mathrm{P} \cdot \mathbf{W_1[:, :r]} \cdot
    \Lambda_{x,z}^B}_{\text{centered in }0 \text{ over }\mathcal{D} \text{ as }\Lambda_{x,z}^\mathcal{D}=0}
    \quad 
    \underbrace{- \quad  \eta \cdot
    \mathrm{P} \cdot \mathbf{W_1[:, r+1:]} \cdot
    \Lambda_{z,z}^B
    }_{\text{always decreases the weights as }\Lambda_{z,z}^B \text{ is PSD} }
    .
\end{split}
\end{equation}
So we have a step that is centered around the step of GD but it is noisy. The noise is given either (1) the first part that is centered around zero over the dataset and (2) the second part that is centered around the GD step.

\paragraph{Two steps: SGD without replacement.}
Eq.\ \ref{eq:GD} implies that if we take a basis of eigenvectors for $P$, $\mathbf{W_1[:, r+1:]}$ components are non-increasing. Moreover, defining $\alpha_{i,j} := P[i,i] \ \Lambda_{z,z}^\mathcal{D} [j,j] \geq 0$, the step on a precise component $\mathbf{W_1[i, j]}$ is given by
$( 1 \ - \eta \alpha_{i,j} ) \cdot \mathbf{W_1[i,j]}.$ When we analyze to steps at time we have essentially 
\[
\Big( 1 \ - \ \eta \underbrace{\left[2 \alpha_{i,j} + \eta \alpha_{i,j}^2 + \eta \cdot \substack{\text{small}\\\text{terms}}\right]}_{\text{denote by }\bar \alpha_{i,j}} \Big) \cdot \mathbf{W_1[i,j]}.
\]
Let us now assume for simplicity for the purpose of this explanation that there are two batches $B_1$ and $B_2$, then define $\alpha,\delta$ such that $\Lambda_{z,z}^{B_1}[j,j] = \Lambda_{z,z}^\mathcal{D}[j,j] - \delta$ and $\Lambda_{z,z}^{B_2}[j,j] = \Lambda_{z,z}^\mathcal{D}[j,j] + \delta$ and analogously that $\mathrm{P}\mathbf{W_1[:, :r]}\Lambda_{x,z}^{B_1}[i,j] = \beta$ and $\mathrm{P}\mathbf{W_1[:, :r]}\Lambda_{x,z}^{B_1}[i,j] = -\beta$.
Here performing two steps on the two batches above as in Eq.\ \ref{eq:SGD} we have some very interesting cancelation that implies a regularization and a speed up in convergence. Indeed two steps together in first approximation can be rewritten as
\[
( 1 \ - \eta \alpha_{i,j} \ - \ \eta^2 \ P[j,j]^2 \delta^2) \cdot \mathbf{W_1[i,j]} \quad + \quad \left[\substack{\text{small terms centered}\\\text{in zero function of }\beta} \right] .
\]

Now, if the depth is higher, denoting the matrix $R = W_L \cdot \ldots \cdot W_3 W_2$, we obtain with GD
\begin{equation}
\begin{split}
    W_1[:, r+1:]^\top
    \quad &\curvearrowleft \quad 
    W_1[:,r+1:]^\top
    \ - \ \eta \cdot
    \E_{D}[z z^\top] W_1[:,r+1:]^\top R^\top R
    .
\end{split}
\end{equation}
Note that both $R^\top R$ and $\E_{D}[z z^\top]$ are symmetric and PSD. This implies that the entries of $W_1[:, r+1:]$ are converging to zero.

In case of mini-batch SGD, even though the expectation over the whole dataset of the multiplication between relevant and irrelevant components is 0, that is not the case on the mini-batch. Thus, assuming the batch are 2 for simplicity, we obtain:
\begin{equation}
\begin{split}
    W_1[:, r+1:]^\top
    \quad &\curvearrowleft \quad 
    W_1[:,r+1:]^\top
    \ - \ \eta \cdot
    \E_{B_1}[z z^\top] W_1[:,r+1:]^\top R^\top R
    \\&\qquad - \ \eta \cdot
    \E_{B_1}[x z^\top] W_1[:,:r]^\top 
    (R - \alpha)^\top R
    .
\end{split}
\end{equation}
The rough idea here is that this implies that after one step we have that the iteration brought the function $Z$ into $Z$ minus $\eta$ that multiplies the following
\begin{equation}
\begin{split}
    \E_{B_1}[z z^\top] Z R^\top R
    \ + \ \E_{B_1}[x z^\top] Z
    (R - \alpha)^\top R,
\end{split}
\end{equation}
and, assuming $\E_{B_2}[xz^\top] = - \E_{B_1}[xz^\top]$, at the second step we have instead that the iteration brought the function $Z$ into $Z$ minus $\eta$ that multiplies the following
\begin{equation}
\begin{split}
    \E_{B_2}[z z^\top] Z R^\top R
    \ - \ \E_{B_1}[x z^\top] Z
    (R - \alpha)^\top R,
\end{split}
\end{equation}
and after two steps, assuming $\E_{B_2}[xz^\top] = - \E_{B_1}[xz^\top]$, we have that the iteration brought $Z$ into into $Z$ plus
\begin{equation}
\begin{split}
    & -2\eta \E_{D}[z z^\top] Z R^\top R 
    \\& - \eta \E_{B_2}[z z^\top] Z (R_1^\top R_1 - R^\top R)
    \\&
    +\eta^2 \E_{B_1}[z z^\top]^2 Z (R^\top R) (R_1^\top R_1)
    \\& + \eta \E_{B_1}[x z^\top] Z
    \big( R_1^\top R_1 - R^\top R  + \alpha^\top (R - R_1) \big)
    \\&
    -\eta^2 \E_{B_1}[x z^\top]\E_{B_1}[z z^\top] Z (R^\top R) (R_1 - \alpha)^\top R_1
    \\&
    +
    \eta^2 \E_{B_2}[z z^\top] \E_{B_1}[x z^\top] Z (R_1^\top R_1) (R - \alpha)^\top R
    \\&
    -
    \eta^2 \E_{B_1}[x z^\top]^2 Z (R-\alpha)^\top R (R_1 - \alpha)^\top R_1
\end{split}
\end{equation}
Let us take now the entry $W_1[j,i]$ of $Z$, let $C = \E_{D}[z z^\top][i,h] = c_i \delta_{i,h}$ and $\E_{B_1}[z z^\top] = C^1$, then the update becomes
\begin{equation}
\begin{split}
    & -2\eta C_i Z[i, :] R^\top R [:, j]
    \\& - \eta C[i, :]^1 Z (R_1^\top R_1 - R^\top R)[:,j]
    \\&
    +\eta^2 C[i, :]^1 (C-C^1) Z (R^\top R) (R_1^\top R_1)[:,j]
    \\& + \eta \E_{B_1}[x z^\top][i,:] Z
    \big( R_1^\top R_1 - R^\top R  + \alpha^\top (R - R_1) \big)[:,j]
    \\&
    -\eta^2 \E_{B_1}[x z^\top][i, :] C^1 Z (R^\top R) (R_1 - \alpha)^\top R_1[:,j]
    \\&
    +
    \eta^2 (C-C^1)[i,:] \E_{B_1}[x z^\top] Z (R_1^\top R_1) (R - \alpha)^\top R[:,j]
    \\&
    -
    \eta^2 \E_{B_1}[x z^\top]^2[i,:] Z (R-\alpha)^\top R (R_1 - \alpha)^\top R_1[:,j]
\end{split}
\end{equation}
and the part that is order 2 in $\eta$ here is
\begin{equation}
\begin{split}
    & - \eta (C-C^1) Z \underbrace{(R_1^\top R_1 - R^\top R)}_{O(\eta)}
    \\&
    -\eta^2 [(C^1)^2-CC^1]  Z (R^\top R)^2
    \\& + \eta \E_{B_1}[x z^\top] Z
    \underbrace{\big( R_1^\top R_1 - R^\top R  + \alpha^\top (R - R_1) \big)}_{O(\eta)}
    \\&
    -\eta^2 (\E_{B_1}[x z^\top] C^1-C^1\E_{B_1}[x z^\top])
    Z (R^\top R) (R - \alpha)^\top R
    \\&
    +
    \eta^2 C \E_{B_1}[x z^\top] Z (R^\top R) (R - \alpha)^\top R
    \\&
    -
    \eta^2 \E_{B_1}[x z^\top]^2 Z ((R-\alpha)^\top R)^2.
\end{split}
\end{equation}
The expectation over sampling batches and over the order in which you see these batches is of this quantity is
\begin{equation}
\begin{split}
    - \quad \eta^2 \mathrm{Var}(C^1)  Z (R^\top R)^2
    \quad - \quad 
    \eta^2 \mathrm{Var}(\E_{B}[x z^\top]) Z ((R-\alpha)^\top R)^2
    \quad + \quad O(\eta^3)
    .
\end{split}
\end{equation}
Note that the variance of these mini-batch expectations is lead by a fourth moment of the input data divided by the batch size. Thus we conclude with the linear scaling rule.

\subsection{Proof of Proposition \ref{prop:2b}, Proposition \ref{prop:SGD}, and Theorem \ref{theo:time}}

\paragraph{Proof of Lemma \ref{lemma:residuals}.} 
We start by proving Lemma \ref{lemma:residuals}.
Note that $\E_{\mathcal{D}}\big[ \epsilon_B \big] $ is exactly the one of one data point divided by the batch size. Let us now prove that $\E_{\x \in \mathcal{D}}\big[ \epsilon_\x \big] \geq \mathcal{R}$ for every linear model defined by a multiplication with $\theta$ we have
\begin{equation}
    \epsilon_\x 
    =
    \theta^\top \x^2 - y \x.
\end{equation}
On the $j-th$ component of the input and output, we have 
\begin{equation}
    \E_{\mathcal{D}}\left[ \epsilon_\x^2 - \E\big[\epsilon_\x\big]^2 \right] 
    \ = \
    \frac{1}{n}\sum_{(\x,y)\in \mathcal{D}} 
    (\theta^\top \x^2 - y\x)^2
    \ - \ \left[
    \frac{1}{n}\sum_{(\x,y)\in \mathcal{D}} 
    \theta^\top \x^2 - y\x
    \right]^2.
\end{equation}
And we obtain that
\begin{equation}
    \min_{\theta}\E_{\mathcal{D}}\left[ \epsilon_\x^2 - \E\big[\epsilon_\x\big]^2 \right] 
    \quad \geq \quad
    \E_{\mathcal{D}}\big[ y^2 \x^2 \big] - \E_{\mathcal{D}}\big[ y \x \big]^2.
\end{equation}
Because the other term is non-negative for every possible $\theta$.

\paragraph{Discussion.}
Exactly as we did for Proposition \ref{prop:GD}, let us restrict WLOG to the subnetwork interacting with an irrelevant component $j$. This is of the form
\[
f_\theta(x)[j] \quad = \quad  \prod_{i=1}^L W_i[j,j] x[j].
\]
Then for all $h$ we have that on a batch $B$ the SGD update on $W_h[j,j]$ is
\[
- \ \eta \left( \prod_{i\neq h } W_i[j,j] \right)^2 \E_{B}[x[j]^2] \cdot W_h[j,j] 
\ + \ 
\eta \left( \prod_{i\neq h } W_i[j,j] \right) \E_{B}[y[j] x[j]].
\]
Consider now the quantity 
\[
G_i \quad := \quad W_{i+1}[j,j]^2 - W_{i}[j,j]^2.
\]
The evolution induced by a step of mini-batch SGD with batch size $B$ on this quantity is
\[
\left[ 1 - \eta^2 \left( \prod_{h\neq i,i+1 } W_h[j,j] \right)^2  \left( 
f_\theta[j] \E_B[x[j]^2] - \E_B[y[j]x[j]] \right)^2
\right] G_i.
\]
Note that these $G_i$ will stop converging when the model will converges. 
More precisely, the shrinking rate of $G_i$ is related to the gradient of the loss, or precisely is the some entry of Hessian of the model times the residuals instead of the gradient of the model times the residuals.
\[
\prod_{h\neq i,i+1 } W_h[j,j] \left[f_\theta[j] \E_B[x[j]^2] - \E_B[y[j]x[j]]\right].
\]
This implies that this converges slowly and do not converge to 0 when this quantity converges fast enough, e.g., when the gradients of the loss converge fast. However, this converges, and converges fast if for some reason the quantity $\prod_{h\neq i,i+1 } W_h[j,j] \left[f_\theta[j] \E_B[x[j]^2] - \E_B[y[j]x[j]]\right]$ will enter in an oscillatory regime.

Next note that with probability one over initialization at every step at most one of the $W_h[j,j]$ is zero, indeed with probability one there exists a strict ordering between them (non of them are exactly the same) and we proved in the proof of Proposition \ref{prop:2} that the smaller get shrunk more. Thus with probability one only one will approach zero. Assume on one step one of them reaches zero. On the next step others do not change and this one gets pushed away from 0 if on the batch used $\E_B[y[j]x[j]] \neq 0$. At this time, with probability 1 all the terms $W_h[j,j]$ are still strongly ordered and since there are at most countable (finite) steps this event has probability 1 until convergence of the algorithms.
Note that this phenomenon not only happens at zero, but happens also as we are approaching zero, precisely, whenever $W_h[j,j]=\epsilon$ and $f_\theta[j] \E_B[x[j]^2] - \E_B[y[j]x[j]] = \delta $ and we have $\epsilon \ll \delta$ we have that at the next step 
\[
W_h[j,j]=\epsilon - \eta \left[ \prod_{i\neq h} W_i[j,j] \right]\delta.
\]

Next assume that the data is misspecified as in the statement of Proposition \ref{prop:SGD}. Then, the average of $f_\theta[j] \E_B[x[j]^2] - \E_B[y[j]x[j]]$ on the data is bigger than $R^2$, and in the case of SGD without replacement these values over the batches average to the whole dataset. Thus, the expectation of the product of terms of the form
\begin{equation}
    \left( 1 - \left[ \prod_{i\neq h} \W_i[j,j] \right]^2 \E_B[x[j]^2] \right) \W_h[j,j]
    \ + \ \E_B[y[j]x[j]] \prod_{i\neq h} \W_i[j,j]
\end{equation}
is the expectation of the product of the following terms, as the order of the batches does not matter for SGD without replacement and they average to zero for SGD with replacement
\begin{equation}
    \left( 1 - \left[ \prod_{i\neq h} \W_i[j,j] \right]^2 \E_B[x[j]^2] \right) \W_h[j,j].
\end{equation}
Now taking the expectation over the batches sampled, for SGD with replacement we obtain \begin{equation}
    \left( 1 - \left[ \prod_{i\neq h} \W_i[j,j] \right]^2 \E_\mathcal{D}[x[j]^2] \right) \W_h[j,j].
\end{equation}
Instead, for SGD without replacement, we obtain a lower value because at every epoch we are multiplying between each other values that are $\E_\mathcal{D}[x[j]^2] + \delta_B$ and the $\delta_B$ deterministically sum up to 0.
The observation that this variable for finite dataset is a multinomial (thus has bounded variance and logarithm) allows us to conclude.


\paragraph{Proof for depth 2.}
Assume we have 2 layers, the quantity $G(t)$ at step (time) $t$ gets multiplied by $(1 -\eta^2 \epsilon^2_{B_t}(t))$ where $\epsilon^2_{B_t}(t)$ is the square of the residuals of the model $W_2[j,j](t)W_1[j,j](t)$ at time $t$ evaluated on the batch $B_t$.
This implies that for all $t$ we have
\[
G(t) \quad = \quad \prod_{i=1}^t \big(1 -\eta^2 \epsilon^2_{B_i}(i)\big) G(0).
\]
Thus that the time $t_\delta$ (number of steps) until $G(t_\delta) = \delta > 0$ is such that
\[
\log(\delta) \quad = \quad \sum_{i=1}^{t_\delta} \log \big(1 -\eta^2 \epsilon^2_{B_i}(i)\big) 
 + \log \big( G(0) \big).
\]
Or analogously that for all $t$ we have
\[
\sum_{i=1}^t \log \big(1 -\eta^2 \epsilon^2_{B_i}(i)\big) 
 = - \log \big( G(0) \big) + \log(\delta_2).
\]
Now we just have to deal with a sum of random variables, for instance, to bound the expectation of the time we can write, if $B_i$ are independently sampled
\[
\E\left[\sum_{i=1}^t \log \big(1 -\eta^2 \epsilon^2_{B_i}(i)\big)\right]
\leq 
\sum_{i=1}^t \log \left(1 -\eta^2 \E\big[\epsilon^2_{B_i}(i)\big] \right)
\sim
- \eta^2 \sum_{i=1}^t \E\big[\epsilon^2_{B_i}(i)\big]
=
- \eta^2 \sum_{i=1}^t \epsilon^2(i)
.
\]
If we apply now Hoeffding inequality on this sum, we have that there exists a constant $C$ such that
\begin{equation}
\begin{split}
    &\mathbb{P}\left[\sum_{i=1}^t \log \big(1 -\eta^2 \epsilon^2_{B_i}(i)\big) + \eta^2 \sum_{i=1}^t \epsilon^2(i) \geq nx\right]
    \\&\leq 
    \mathbb{P}\left[\sum_{i=1}^t \log \big(1 -\eta^2 \epsilon^2_{B_i}(i)\big) - \eta^2 \sum_{i=1}^t \epsilon^2(i) - nx \geq 0\right]
    \\&\leq 
    \prod_{i=1}^t \E\left[ (1 -\eta^2 \epsilon^2_{B_i}(i))^\lambda
    \right] 
    \E_\mathcal{D}\left[\exp\left(-\eta^2 \lambda \sum_{i=1}^t \epsilon^2(i)\right)\right]
    \exp(-\lambda t x)
    \\&\leq 
    \left[ \substack{\text{upper bound to}\\ \text{second order Taylor}\\
    \text{of exponential}}  = C t \lambda^2\right]
    \exp(-\lambda t x)
    \\&\leq
    \exp\left( -\frac{2 t x}{C} \right).
\end{split}
\end{equation}
Thus with probability bigger than $1 - \delta_1$ we have that $\mednorm{G_t}_F = \delta_2$ 
when $t$ is such that
\[
\sum_{i=1}^t \epsilon^2(i)
= O\left(
\frac{\log\big( \mednorm{G_0}_F \big) - \log(\delta_2) - \log(\delta_1)}{\eta^2} 
\right)
.
\]
Here if we assume that the series of the $\epsilon$ does not converge, we obtain the thesis. More in general, whenever there exists a $t$ such that the series
\[
\sum_{i=1}^t \E\big[\epsilon^2_{B_i}(i)\big]
\quad \geq \quad \frac{\log\big( \mednorm{G_0}_F \big)}{\eta^2}
\]
We have convergence, this proves Proposition \ref{prop:2b} of two layers (or of many layers but about the smallest two) in the case of SGD with replacement.
In the case of SGD without replacement we have, by applying the lemma on expectations of this form in \cite[Lemma 12 or Corollary 13]{beneventano_trajectories_2023}, that 
\[
\E\left[ \prod_{i=1}^t (1 -\eta^2 \epsilon^2_{B_i}(i))^\lambda
    \right] 
    <
\prod_{i=1}^t \E\left[ (1 -\eta^2 \epsilon^2_{B_i}(i))^\lambda
    \right] 
\]
instead of being equal, thus the bound could be even stronger and we conclude that in the case of SGD without replacement convergence is faster (although we can not quantify it).

To now conclude the proof of Proposition \ref{prop:SGD} and Theorem \ref{theo:time} we just need to apply Lemma \ref{lemma:residuals}.
Then reapplying Hoeffding inequality to the $\epsilon$s, we obtain that with probability
bigger than $1 - \delta_1$ we have that $\mednorm{G_t}_F = \delta_2$ 
when $t$ is such that
\begin{equation}
\label{eq:2-layers}
t 
= O\left(
\frac{b}{\eta^2 \mathcal{R}_j^2} \left( \log\big( \mednorm{G_0}_F \big) - \log(\delta_2) - \log(\delta_1) \right)
\right)
.
\end{equation}

\paragraph{The case of higher depth.}
WLOG, on the $j$th component, we are in the case in which we are optimizing the following
\[
\frac{1}{n} \sum_{(\x,y) \in \mathcal{D}} \Big\|\prod_{i = 1}^D \theta_i \x[j] - y[j]\Big\|^2
\]
with $\theta_1, \theta_2, \ldots, \theta_D \in \R$ and $|\theta_1| < |\theta_2| < \ldots < |\theta_D|$. Since the $\theta_i$ are numbers we can do commutation WLOG and the union of the events in which two of them are equal at any point during the training is a union of numerable events of probability 0, thus has probability 0.

Analogously to what we did before, $\theta_1$ will converge to 0 exponentially in time $O(\eta^{-1})$.

Consider now the following quantities $G_i^j \in \R$ defined as
\[
G_i^1 = \theta_i, \qquad G_i^2 = \theta_i^2 - \theta_1^2
\]
and 
\[
G_i^j \quad = \quad \big(G_i^{j-1}\big)^2 \ - \ \big(G_{j-1}^{j-1}\big)^2.
\]
We will prove here by induction the following lemma
\begin{lemma}
\label{lemma:2}
    Assume that for all possible models $f_\theta(x)$ and the process of sampling batches we have $\E_{\x \in \mathcal{D}}\Big[ \E_{\x \in B}\big[ \epsilon_\x\big]^2 \Big] = c > 0$ lower bounded by $c>0$. Assume that $0<\eta<2/\lambda_{\max}$ where $\lambda_{\max}$ is the highest eigenvalue of the Hessian at initialization.
    Then $G_j^j(t)$ converges to zero exponentially fast with rate $\eta^{2^{j-1}}c^{2^{j-1}}$.
    More precisely, let $\delta_1, \delta_2 > 0$ small, then with probability higher than $1-\delta_2$, we have $G_j^j(t) \leq \delta_1^{2^{j}}$ and $\theta_j(t) \leq \delta_1$ in number of steps that is
    \begin{equation}
    \label{eq:i-layers}
    t \quad = \quad
    O\left( 
    \frac{1}{\eta^{2^{j-1}} c^{2^{j-1}}} \big( 2^{j-1}\log(\theta_j(0)) - 2^{j}\log(\delta_1) - \log(\delta_2)\big)
    \right)
\end{equation}
for both SGD with and without replacement but for SGD without replacement it is strictly faster.
\end{lemma}

We already proved it in the case of $G_i^2$ above for $\theta_2$. It will converge to 0 exponentially in time $O(\eta^{-2})$, and we will see now analogously the inductive step.

Recall we defined $\epsilon_\x = f_\theta(\x)\x^\top - y\x^\top$ for $\x \in \mathcal{D}$. In this proof, since we are working on the $j$th component $\epsilon_\x = \prod_{i=1}^D \theta_i \x[j]^2 - y[j]\x[j]$.
Let us first write down the dynamics for the system of the $G_j$. We have for all $i$ that one step of SGD with batch $B$ and step size $\eta>0$ is
\begin{equation}
\begin{split}
    G_i^1(t + 1) 
    \quad = \quad 
    G_i^1(t) - 
    \eta \E_{B}\big[\epsilon_\x\big] \prod_{j \neq i} G_j^1(t).
\end{split}
\end{equation}
Next note that, as we showed above
\begin{equation}
\begin{split}
    G_i^2(t + 1) 
    \quad &= \quad 
    G_i^2(t) - 
    \eta^2 \E_{B}\big[\epsilon_\x\big]^2 \prod_{j \neq i,1} \theta_j(t)^2
    \\&= \quad 
    G_i^2(t) - 
    \eta^2 \E_{B}\big[\epsilon_\x\big]^2 \prod_{j \neq i,1} \big(G_j^1(t)\big)^2 \ + \  O(\eta^2 \E_{B}\big[\epsilon_\x\big]\theta_1^2).
    \\&= \quad 
    G_i^2(t) - 
    \eta^2 \E_{B}\big[\epsilon_\x\big]^2 \prod_{j \neq i} G_j^2(t) \ + \  O(\eta^2 c^2 \delta_1^2)
\end{split}
\end{equation}
since $|\theta_1| < \delta_1 < \eta$.
Analogously, by induction we have that on step of SGD on $G_i^h$ is
\begin{equation}
\begin{split}
    G_i^j(t + 1) 
    \quad &= \quad 
    \big(G_i^{j-1}(t + 1) \big)^2 - \big(G_{j-1}^{j-1}(t + 1) \big)^2
    \\&= \quad 
    G_i^j(t) - \big(\eta^{2^{j-2}}\big)^2  
    \left(\E_{B}\big[\epsilon_\x\big]^{2^{j-2}}\right)^2 \prod_{\substack{h \neq i\\ h \geq j }} \big(G_h^{j-1}(t)\big)^2 
    \\ &\quad + \quad  2\eta^{2^{j-2}}\E_{B}\big[\epsilon_\x\big]^{2^{j-2}}\prod_{\substack{h \neq i\\ h \geq j }} G_h^{j-1}(t) \ \cdot \ \left[\substack{\text{rest of}\\j-1}\right]
    \\&= \quad 
    G_i^j(t) - \big(\eta^{2^{j-2}}\big)^2  
    \left(\E_{B}\big[\epsilon_\x\big]^{2^{j-2}}\right)^2 \prod_{\substack{h \neq i\\ h \geq j }} \big(G_h^{j-1}(t)\big)^2 
    \ + \  O\left( \eta c \delta_1 \right)^{2^{j-1}}
    \\&= \quad 
    G_i^j(t) -
    \eta^{2^{j-1}} \E_{B}\big[\epsilon_\x\big]^{2^{j-1}} \prod_{\substack{h \neq i\\ h \geq j }} G_h^j(t) \quad + \quad  O\left( \eta c \delta_1 \right)^{2^{j-1}}.
\end{split}
\end{equation}

So we just proved that approximately, the dynamical system of the $G^j$ for all $j$s is the same.
Next note that thus, if $G^j_i(t) < G^j_h(t)$ for some $i,j,h,t$ then the same holds for all $t$.
Thus the case for $j=2$ proves it for all $j$. This concludes the proof of Lemma 2.

\subsection{Proof of Proposition \ref{prop:3} and Proposition \ref{prop:3a}}

The $u$ components that are not spanned by the data are untouched by the SGD or GD step. This implies that the solve the following recursion $X_0=W_1[i,j]$ at initialization
\[
X_{k+1} \quad = \quad (1 - \eta \lambda) X_k.
\]
Thus if $\eta < 2/\lambda$ the convergence is exponential.
This concludes already the proof of Proposition \ref{prop:3a}.

Regarding the $i$ irrelevant components that are spanned by the data for the linear model, in the notations of Eq.\ \ref{eq:GD_decreases_proof}, GD plus weight decay on the components of $\bar W_1[i, j]$ is
$W_1[:,r+1:]$ is
\begin{equation}
\begin{split}
    (1 - \eta D_W[i,i] D_x[j,j] - \eta \lambda) \bar W_1[i, j].
    .
\end{split}
\end{equation}
In particular, the rate of decrease is lower bounded by $(1 - \eta \lambda)$ thus we have exponential convergence to 0.
Moreover, the speed at every step is
$(1 - \eta \lambda - \text{GD part})$ for GD and 
$(1 - \eta \lambda - \text{GD part} - \text{SGD part})$ in the case of SGD. Thus SGD + weight decay is faster.
This concludes the proof of Proposition \ref{prop:3} and Proposition \ref{prop:need4wd}.


\end{document}